\documentclass[3p]{elsarticle}
\journal{International Journal of Approximate Reasoning}

\usepackage{etoolbox}
\makeatletter
\patchcmd{\ps@pprintTitle}
  {Preprint submitted}
  {Accepted for publication}
  {}{}
\makeatother

\usepackage[linesnumbered,ruled,vlined,plain,norelsize]{algorithm2e}
\usepackage{tikz}
\usepackage{sistyle}
\usepackage{longtable}
\usepackage{graphicx}
\usepackage{threeparttable}
\usepackage{amsmath, amsfonts, amssymb, amsthm}
\usepackage{mathtools}
\usepackage{array}
\usepackage{enumitem}
\usepackage{booktabs}

\DeclareMathOperator*{\drc}{\scalerel*{\circled{$\cup$}}{\textstyle\sum}}
\DeclareMathOperator*{\crc}{\scalerel*{\circled{$\cap$}}{\textstyle\sum}}

\usepackage{arev}
\DeclareSymbolFont{extraup}{U}{zavm}{m}{n}
\DeclareMathSymbol{\varheart}{\mathalpha}{extraup}{86}

\usepackage{scalerel}
\usepackage{mathtools}
\usepackage{bm}
\usepackage{array}
\usepackage{color, colortbl}
\usepackage[framemethod=default]{mdframed}
\usepackage{footnote}
\usepackage{tablefootnote}
\usepackage{gnuplottex} 
\usepackage{subcaption}
\usepackage{pgfplots}
\usepgfplotslibrary{colormaps}
\pgfplotsset{
    colormap={slategrayred}{
		rgb255=(255,129,129)
		rgb255=(215,215,0)		
		rgb255=(140,255,140)
	},
    colormap={slategrayred1}{
		rgb255=(255,129,129)
		rgb255=(140,255,140)
	},
    colormap={slategraygreen}{
		rgb255=(0,255, 0)
        rgb255=(0,255, 0)
	},
    colormap={slategraymark}{
        rgb255=(255,102,102)
		rgb255=(198,255,179)
	},
    colormap={slategray}{
		rgb255=(105,105, 255)
        rgb255=(105,105, 255)
	},	
}
\usepackage{xcolor}
\usepackage{empheq}
\usepackage{hyperref}
\usepackage{multirow}
\usepackage{makecell}
\usepackage{filecontents}
\usepackage{dsfont}
\usepackage{pgfplots}
\usepgfplotslibrary{groupplots}
\pgfplotsset{compat=newest}

\usepgfplotslibrary{external} 
\makeatletter
\newcommand{\removelatexerror}{\let\@latex@error\@gobble}

\newcommand{\docircint}[2]{%
	\ifx#1\displaystyle
	\displaycircint
	\else
	\normalcircint{#1}%
	\fi
}
\newcommand{\displaycircint}{\displaystyle\mathsf{C}\mkern-18mu}
\newcommand{\normalcircint}[1]{%
	\smallerc{#1}\ifx#1\textstyle\mkern-9mu\else\mkern-8.2mu\fi
}
\newcommand{\smallerc}[1]{%
	\vcenter{\hbox{$\ifx#1\textstyle\scriptstyle\else\scriptscriptstyle\fi\mathsf{C}$}}%
}

\newcommand{\docircintmin}[2]{%
	\ifx#1\displaystyle
	\displaycircintmin
	\else
	\normalcircintmin{#1}%
	\fi
}
\newcommand{\displaycircintmin}{\displaystyle\mathsf{c}\mkern-10mu}
\newcommand{\normalcircintmin}[1]{%
	\smallercmin{#1}\ifx#1\textstyle\mkern-6mu\else\mkern-5.2mu\fi
}
\newcommand{\smallercmin}[1]{%
	\vcenter{\hbox{$\ifx#1\textstyle\scriptstyle\else\scriptscriptstyle\fi\mathsf{c}$}}%
}

\newsavebox\myboxA
\newsavebox\myboxB
\newlength\mylenA

\newcommand*\xoverline[2][0.75]{%
	\sbox{\myboxA}{$\m@th#2$}%
	\setbox\myboxB\null
	\ht\myboxB=\ht\myboxA%
	\dp\myboxB=\dp\myboxA%
	\wd\myboxB=#1\wd\myboxA
	\sbox\myboxB{$\m@th\overline{\copy\myboxB}$}
	\setlength\mylenA{\the\wd\myboxA}
	\addtolength\mylenA{-\the\wd\myboxB}%
	\ifdim\wd\myboxB<\wd\myboxA%
	\rlap{\hskip 0.5\mylenA\usebox\myboxB}{\usebox\myboxA}%
	\else
	\hskip -0.5\mylenA\rlap{\usebox\myboxA}{\hskip 0.5\mylenA\usebox\myboxB}%
	\fi}

\DeclareRobustCommand\bigop[1]{%
	\mathop{\vphantom{\sum}\mathpalette\bigop@{#1}}\slimits@
}
\newcommand{\bigop@}[2]{%
	\vcenter{%
		\sbox\z@{$#1\sum$}%
		\hbox{\resizebox{\ifx#1\displaystyle.9\fi\dimexpr\ht\z@+\dp\z@}{!}{$\m@th#2$}}%
	}%
}

\newcommand*\circled[1]{\tikz[baseline=(char.base)]{
		\node[shape=circle,draw,inner sep=2pt] (char) {#1};}}
\makeatother

\newenvironment{cons}{\fontfamily{pcr}\selectfont}{\par}
\DeclareTextFontCommand{\textmyfont}{\cons}

\theoremstyle{plain}

\newtheorem{theorem}{Theorem}
\newtheorem{definition}[theorem]{Definition}
\newtheorem{note}{Note}

\begin{document}
	
\usetikzlibrary{calc, arrows, automata, positioning, patterns}
\tikzexternaldisable

\begin{frontmatter}
\title{Effectiveness Assessment of Cyber-Physical Systems}

\author[add1,add2]{G\'erald Rocher}
\ead{gerald.rocher@gfi.fr}

\author[add2,add3]{Jean-Yves Tigli}
\ead{jean-yves.tigli@unice.fr}

\author[add2,add3]{St\'ephane Lavirotte}
\ead{stephane.lavirotte@unice.fr}

\author[add2,add3]{Nhan Le Thanh}
\ead{nhan.le-thanh@unice.fr}

\address[add1]{GFI Informatique, Saint-Ouen, France}
\address[add2]{Universit\'e C\^ote d'Azur (UCA), Sophia Antipolis, France}
\address[add3]{CNRS, laboratoire I3S, Sophia Antipolis, France}

\begin{abstract}
By achieving their purposes through interactions with the physical world, Cyber-Physical Systems (CPS) pose new challenges in terms of dependability. Indeed, the evolution of the physical systems they control with transducers can be affected by surrounding physical processes over which they have no control and which may potentially hamper the achievement of their purposes. While it is illusory to hope for a comprehensive model of the physical environment at design time to anticipate and remove faults that may occur once these systems are deployed, it becomes necessary to evaluate their degree of effectiveness in vivo. In this paper, the degree of effectiveness is formally defined and generalized in the context of the measure theory. The measure is developed in the context of the Transferable Belief Model (TBM), an elaboration on the Dempster-Shafer Theory (DST) of evidence so as to handle epistemic and aleatory uncertainties respectively pertaining the users' expectations and the natural variability of the physical environment. The TBM is used in conjunction with the Input/Output Hidden Markov Modeling framework we denote by Ev-IOHMM to specify the expected evolution of the physical system controlled by the CPS and the tolerances towards uncertainties. The measure of effectiveness is then obtained from the forward algorithm, leveraging the conflict entailed by the successive combinations of the beliefs obtained from observations of the physical system and the beliefs corresponding to its expected evolution. The proposed approach is applied to autonomous vehicles and shows how the degree of effectiveness can be used for bench-marking their controller relative to the highway code speed limitations and passengers' well-being constraints, both modeled through an Ev-IOHMM.
\end{abstract}

\begin{keyword}
	Cyber Physical Systems, Degree of Effectiveness, Transferable Belief Model, Input/Output Hidden Markov Model, Zone of Viability
\end{keyword}

\end{frontmatter}

\captionsetup{justification=centering}

\SetKwData{x}{x}

\section{Introduction}
\label{sec:introduction}
\noindent Generally, computing systems are understood as being purposeful processing units, directed to produce expected results by means of computational resources manipulating data through controlled computational environments.\medskip

\noindent At the infrastructure level, some hardware and software mechanisms ensure correct operation of the computing resources (e.g. power-on self-test, etc.), integrity and persistence of the data (e.g. Cyclic Redundancy Check (CRC), memory content refresh, etc.). At the system level, accesses to the computational resources are made safe by an operating system or a middleware. The computational environments being controlled, the production and the persistence of the expected results are guaranteed "by design" solely provided that the computer program issues the right commands to the computational resources. In this sense, a computer program is a \textit{perfect deterministic model} of a computing system and the question does not even arise that, \x being a variable, the execution of the following code snippet will lead its value to be set to 6 into the memory. 

\begin{figure}[!ht]
\setlength{\interspacetitleruled}{0pt}
\setlength{\algotitleheightrule}{0pt}
	\removelatexerror
	\begin{algorithm}[H]		
		\SetKwFunction{Wait}{Wait}
		\SetKwFunction{Add}{Add}
		\SetKwFunction{Assert}{Assert}		
		$\x \gets 1$\;
		\Wait{$10000s$}\;
		\Add{$\x,5$}\;
		\Wait{$10000s$}\;
		\Assert{$\x,6$}\tcp*{TRUE}
	\end{algorithm} 
\end{figure}

\noindent Let us now consider Cyber-Physical Systems (CPS) as being orchestrations of distributed computing and physical systems \cite{lee2015past}. CPS can be understood as being \textit{"cyber" physical processes} where some properties of a physical system of interest are purposefully modified by means of computational resources manipulating them through transducers (e.g. sensors and actuators). For instance, let us keep the template of the preceding code snippet by considering that the variable to be modified now corresponds to a physical property of the physical system (e.g. the temperature in a living room).

\begin{figure}[!ht]
\setlength{\interspacetitleruled}{0pt}%
\setlength{\algotitleheightrule}{0pt}%
	\removelatexerror
	\begin{algorithm}[H]
		\SetKwFunction{Wait}{Wait}
		\SetKwFunction{SetTemperature}{SetTemperature}
		\SetKwFunction{IncreaseTemperature}{IncreaseTemperature}
		\SetKwFunction{Assert}{Assert}
		\SetKwData{LivingRoom}{LivingRoom}
		\SetKwData{Temperature}{Temperature}		
		\SetTemperature{\LivingRoom, \SI{18}{\degC}}\;
		\Wait{$10000s$}\;
		\IncreaseTemperature{\LivingRoom, \SI{5}{\degC}}\;
		\Wait{$10000s$}\;
		\Assert{\Temperature, \LivingRoom, \SI{23}{\degC}}\tcp*{???}
	\end{algorithm} 
\end{figure}
\medskip
\noindent What trust can we have that the temperature in the living room is going to be changed to \SI{23}{\degC}? In other words, can one consider the above code snippet as a perfect deterministic model of the physical system? Considering that the living room is a \textit{non-isolated} physical system, the answer is "no". Such systems are driven by \textit{non-deterministic dynamics}, at any time, the temperature of the living room can be affected by surrounding processes over which the computing system has no control \cite{garlan2010software} \cite{bures2015software} \cite{zhang2016understanding}. This situation is aggravated for the Internet of Things (IoT)-based CPS whose underlying infrastructure is volatile. Indeed, their structural components being embedded into physical things, their availability cannot be ensured over time. Consequently, the attainment of the CPS purposes cannot be guaranteed solely "by design" \cite{lee2015past}.

\medskip 
\noindent As a solution to this problem, we propose to quantitatively assess, at run-time, to which extent the CPS purposes are met. In other words, it is about providing the \textit{degree of effectiveness} of the CPS as a \textit{measure} of the concrete evolution of the physical system according to the expected evolution. To be more precise about the measure and the meaning we seek to give it as an assessment of the degree of effectiveness of the CPS, we borrow some terminology employed in the viability theory \cite{aubin:inria-00636570}. Let us assume that the expected evolution of the physical system can be specified as a deterministic model, free from \textit{uncertainties}, where (1) state transitions are determined by contextual events (stimuli), (2) states are qualified by the expected physical effects resulting from actuators over which the computing system has control. \textit{Zones of Viability} extend this deterministic point of view with tolerances accounting for aleatory uncertainties pertaining the natural variability of actuators effects and sensors readings and for epistemic uncertainties relative to the users' satisfaction towards the concrete evolution of the physical system.\\
\noindent In this paper, we propose to generalize the deterministic model in the framework of the measure theory. Doing so, one can leverage the set of measures (probabilities, possibilities, etc.) as a means of defining zones of viability from which one can reason in order to obtain the degree of effectiveness. By obtaining a quantitative measure of the degree of effectiveness, (1) one can leverage this measure within a feedback loop so as for the controller of the system to minimize the behavioral drift (e.g. negative feedback control systems \cite{bellman2015adaptive}), (2) one can use this measure as a bench-marking tool used to compare algorithms deployed for controlling CPS.

\section{Related work and contributions}
\label{sec:related_work}
\noindent The work presented in this paper is closely related to the \textit{dependability} of the computing systems \cite{laprie1992dependability}. Within computer science, this term refers to the trust that can justifiably be placed in the service delivered by computing systems and covers all their critical quality aspects \cite{eusgeld2008dependability}. In other words, it reflects users' degree of trust in these systems. Among the attributes of dependability \cite{avizienis2004basic}, \textit{availability} (i.e. readiness for correct service), \textit{reliability} (i.e. continuity of correct service), \textit{safety} (i.e. absence of consequences on the users and the environment) and \textit{integrity} (i.e. absence of improper system alterations) characterize the immunity of computing systems towards uncontrolled physical processes and associated uncertainties (i.e. \textit{threats} that can affect computing systems operation and undermine their dependability \cite{avizienis2004basic}).\\
\noindent The assessment of the dependability can be done at design time through analytic metrics using models of the systems and, whenever possible, the known uncertainties (e.g. U-Test \cite{utest}). Run-time monitoring involves \textit{direct} and \textit{indirect} \textit{empirical} metrics, respectively measuring the system itself through probes (whenever possible) and its effects within the physical environment through sensors.\\
\smallskip

\noindent While methodologies involved at design time (e.g. Model-based design) and at testing phase (e.g. Model checking, simulation, etc.) are respectively devoted to \textit{fault prevention} and \textit{fault removal}, run-time monitoring is devoted to automatic \textit{fault and anomaly detection} \cite{delgado2004taxonomy}. The most common formulation of the anomaly detection problem is to determine if a given test sequence is anomalous with respect to normal sequences. More formally, given a set of $n$ normal sequences $S=\{(S_k)_{1,k\in\mathbb{N}^\star},\ldots,(S_k)_{n,k\in\mathbb{N}^\star}\}$ and a test sequence $(\mathfrak{S}_k)_{k\in\mathbb{N}^\star}$, it is about computing an anomaly \textit{score} for $(\mathfrak{S}_k)_{k\in\mathbb{N}^\star}$, with respect to $S$. It is assumed that test sequences might be misaligned in time and space \textit{w.r.t} the normal sequences. We do also consider \textit{complex}, and \textit{collective anomalies}. On the one hand, when contextual attributes can be associated with observations (e.g. time, location, etc.), contextual anomalies are corresponding to behaviors that are valid under some conditions but are abnormal in others. For instance, in European countries, normally high temperatures during the summer can be considered as contextual anomalies if they occur during the winter (time-based contextual anomaly). On the other hand, collective anomalies correspond to a collection of consecutive behaviors which are not abnormal by themselves but are abnormal when they occur together as a collection \cite{chandola2009anomaly}. Approaches that address these anomalies fall into three categories described hereafter.
\subsection{Prediction-based approaches}
\noindent \textbf{These approaches consist in modeling legitimate behavior through a parametric model learned from observations and further used for predicting observation at each time $t$. Abnormal behaviors are those whose real observations differ from the predicted ones.}\\

\noindent In \cite{malhotra2015long}, authors use stacked Long Short-Term Memory (LSTM) networks \cite{hochreiter1997long} for anomaly/fault detection in time series. A network is trained on non-anomalous data and used as a predictor over a number of time steps. The resulting prediction errors are modeled as a multivariate Gaussian distribution, which is used to assess the likelihood of anomalous behavior. In \cite{goh2017anomaly}, authors present an unsupervised approach to detect cyber-attacks in Cyber-Physical Systems (CPS). A Recurrent Neural Network (RNN) \cite{mandic2001recurrent} is used as a time series predictor. The Cumulative Sum method is further used to identify anomalies in a replicate of a water treatment plant.\\
\medskip

\noindent \textbf{pros \& cons:} these models are difficult to train \cite{pascanu2013difficulty} and are generally hardly interpretable, their intrinsic structure and parameters making unclear the mapping between the variables and the observations \cite{lipton2016mythos}. For instance, such models, once learned make difficult, if not impossible, the modification of their intrinsic parameters in order to tune a posteriori the tolerances pertaining the epistemic uncertainties. More importantly, learning a comprehensive model of the CPS behavior based on observations is often impracticable with regards to their complexity \cite{lee2015past}.
\subsection{Model drift-based approaches}
\noindent \textbf{These approaches are relative to the anomalous evolution of the model parameters. The basic idea is to build a parametric behavioral test model from test sequences as they arrive and compare it with the normal behavioral model. Dissimilarities between models give the anomaly score.}\\

\noindent Authors in  \cite{webb2016characterizing} focus on the quantitative measure of \textit{concept drift} and introduce the notion of drift magnitude whose value can be quantified through distance functions such as Kullback-Leibler Divergence or Hellinger Distance. Close to the idea of concept drift is the notion of \textit{Bayesian Surprise} \cite{itti2009bayesian}. A surprise quantifies how data affects an observer. It quantifies a mismatch between an expectation and what is actually observed by measuring the difference between posterior and prior beliefs of the observer. In \cite{storck1995reinforcement} authors propose using Bayesian surprise as a measure of the learning progress of reinforcement learning agents.\\
\smallskip

\noindent \textbf{pros \& cons:} being based on the distance between prior and posterior beliefs, the main disadvantage of these approaches concerns the speed of convergence to an accurate test model, highly dependent on the number of observations needed to learn it. Hereby, a short time anomalous behavior might be "attenuated" or even not detected. These approaches are mainly leveraged in autonomic computing and the models@run-time community \cite{assmann2012models} where an initial model is updated over time taking into account unanticipated evolutions of the environment. In this context, above a given threshold, the quantitative drift value is used to trigger the update of the model with the newly learned parameters, assuming it represents the correct behavior.
\subsection{Likelihood-based approaches}
\noindent\textbf{These approaches consist in modeling legitimate behavior through a parametric model and considering abnormal behaviors as those having low "likelihood" to have been generated by the model.}\\

\noindent In this category, Dynamic Bayesian networks (DBN) and derivatives ($n$-order Markov chains) are widely used where tolerances towards uncertainties are generally described through probability density functions (pdf). An extension of the Markovian models, denoted by Hidden Markov Models (HMM), consists in considering the case where states of the model are "hidden" \cite{rabiner1989tutorial}, i.e. not directly observable, or partially hidden \cite{ramasso2014making}. Such models are particularly well suited in the context of this paper where it is assumed that while the expected behavior of a CPS can be described a priori, the prior knowledge of its concrete internals and surrounding environment is unlikely available \cite{lee2015past}. In this context, the likelihood of a given observation sequence $(\vec{y}_t)_{t=1}^T$ is inferred from the model of the expected behavior of the system by using the probabilistic forward algorithm. This algorithm computes the likelihood of all the possible sequences of hidden states given the observation sequence $(\vec{y}_t)_{t=1}^T$. The likelihood of a particular sequence of hidden states $(x_t)_{t=1}^T$ given the observation sequence $(\vec{y}_t)_{t=1}^T$ is given by:
\begin{equation}
		p\left((x_t)_{t=1}^T | (\vec{y}_t)_{t=1}^T\right)=p(x_1)\times p(x_1|\vec{y}_1)\times\left[\prod_{t=2}^T\left(p(x_{(t)}|x_{(t-1)})\times p(x_{(t)}|\vec{y}_{(t)})\right)\right]
\end{equation}
\noindent Some works have extended the HMM in the framework of the Transferable Belief Model (TBM) \cite{smets1994transferable}, an elaboration on the Dempster-Shafer Theory (DST) of evidence where tolerances towards uncertainties are neither described by probabilities but by belief functions. In \cite{ramasso2017inference}, the author describes previous works in using HMM with TBM \cite{ramasso2007forward},\cite{ramasso2007reconnaissance} in the context of analyzing time series and denoted as Evidential HMM (EvHMM). Probability-based HMM is built upon the Closed World Assumption (CWA), i.e. probabilities are spread on the states $\{x_1,\ldots,x_N\}$ defined in the model with $\sum_{i=1}^N p(x_i)=1$, i.e. $p(\Omega)=1$ and $p(\emptyset)=0$. TBM, on his side, is built upon the Open World Assumption (OWA). It allows to associate a belief value (mass of conflict $m$) to the empty set, i.e. $m(\emptyset)\geq 0$, meant to quantify the degree of inconsistency of the observations with regards to the model. This is coherent with the meaning we seek to give to the measure of effectiveness. In this context, it is proven in \cite{smets2002application} that the \textit{plausibility} of the observation sequence to have been produced by the model, i.e. the plausibility of the model, is given by $pl(\Omega)=1-m(\emptyset)$ obtained from the evidential forward algorithm, likewise the likelihood obtained from the probabilistic forward algorithm.\\
\noindent Close to the concern of CPS behavioral analysis, the case of Evidential HMM with application to dynamical system analysis is described in \cite{serir2011time}. However, HMM-based methods do not consider state-transitions probabilities governed by inputs necessary in modeling CPS expected behavior.\\
\noindent A way to cope with this limitation is to use the Input/Output HMM (IOHMM), first introduced in \cite{bengio1995input}. With this modeling framework, state-transitions probabilities are not hardcoded as it is the case with HMMs. Instead, the probability of a state-transition to occur depends on some input values. In this context, the observation sequence consists in an input sequence $(\vec{u}_t)_{t=1}^T$ and an output sequence $(\vec{y}_t)_{t=1}^T$. In this context, the likelihood of a particular sequence of hidden states $(x_t)_{t=1}^T$ given the sequences $(\vec{u}_t)_{t=1}^T$ and $(\vec{y}_t)_{t=1}^T$ is given by :  

\begin{equation}
	p\left((x_t)_{t=1}^T,(\vec{u}_t,\vec{y}_t)_{t=1}^T\right)=p(x_1)\times p(x_1|\vec{y}_1)\times\left[\prod_{t=2}^T\left(p(x_{(t)}|x_{(t-1)},\vec{u}_{(t-1)})\times p(x_{(t)}|\vec{y}_{(t)})\right)\right]
\end{equation}

\noindent \textbf{pros \& cons:} a key advantage here is that these models are interpretable, making clear (1) the mapping between the variables and the observations, (2) the description of the zones of viability through probabilities or belief functions. In this category, HMM-based modeling frameworks and more particularly the IOHMM where state-transitions probabilities depend on some input values, are well suited for representing dynamical systems \cite{al2011hidden}\cite{fraser2008hidden}. Moreover, such models assume that the internals and the environment of the systems considered are not necessarily known a priori. This makes sense in the context of CPS that, with regards to their complexity, are unlikely to be comprehensively modeled. At best, one can define their expected behavior through the effects they are supposed to produce in response to some events. \textbf{By assuming OWA, the Ev-IOHMM would be a good candidate so as to compute the degree of effectiveness of CPS. However, to date, no effort has been put on elaborating on such modeling framework}. 

\subsection{Contributions}
\noindent In this paper, we do extend previous works on the probabilistic and the possibilistic IOHMM likelihood-based approaches respectively described in \cite{rocher2017probabilistic} and \cite{rocher2018possibilistic} in the framework of the TBM (we denote Ev-IOHMM). The work done in \cite{ramasso2009contribution},\cite{smets2008belief} and \cite{shenoy1992valuation} being considered as the starting points, the main contributions of this paper are the following:
\begin{enumerate}
	\item The degree of effectiveness is formally defined and generalized in the context of the measure and the viability theories,
	\item The probabilistic IOHMM described in \cite{bengio1995input} is extended into the TBM framework, resulting in the Ev-IOHMM. To this end, we do rely on previous contributions done on extending HMM to EvHMM \cite{smets2002application}\cite{ramasso2007forward}. The associated evidential forward algorithm is provided and used for inferring the likelihood of the input/output observations to have been generated by the model whose zones of viability are neither defined through probabilities \cite{rocher2017probabilistic} nor possibilities \cite{rocher2018possibilistic} but by belief functions.
	\item The Evaluation of the approach is carried out on a simple yet revealing example, complemented with a list of use-cases emphasizing its interest. Among these use-cases, we do elaborate on a use-case in the domain of autonomous vehicles. The idea is to leverage the proposed approach as means for designers to benchmark the control systems of these vehicles by relying on the measure of their effectiveness against constraints of the highway code relative to speed limitations and passengers well-being, both modeled through an Ev-IOHMM.   
\end{enumerate}

\SetKwData{low}{lowLuminosity}
\SetKwData{high}{highLuminosity}
\SetKwData{p}{pres}
\SetKwData{l}{lum}
\SetKwData{a}{AND}
\SetKwData{o}{OR}
\SetKwData{v}{Viab}
\SetKwData{cf}{Cfrt}
\SetKwData{u}{Tol}

\section{Mathematical background}
\subsection{Deterministic model of the expected behavior}
\label{sec:dfsa}
\noindent In this paper, we do consider physical systems whose expected evolution under a CPS control can be \textit{constrained} through a deterministic model $\lambda$ whose state transitions are determined by contextual events (stimuli) while states are qualified by the expected physical effects resulting from actuators over which the computing system has control. This model is formally defined by:
\begin{equation}
\left\{
\begin{array}{l l l}
x_{(t)} &= \Gamma(x_{(t-1)},\vec{u}_{(t-1)}) & \text{(\textbf{State prediction})}\\
\psi_{(t)} &= G(x_{(t)}) & \text{(\textbf{State emission})}\\

U_{(t)}&= \mathcal{U}(x_{(t-1)},x_{(t)}) & \text{(Finite set of inputs)}\\
Y_{(t)}&= \mathcal{Y}(x_{(t)}) & \text{(Finite set of outputs)}
\end{array}
\right.
\label{_dfsa}
\end{equation}

\noindent with:
\begin{itemize}
	\item[--] $t\in\mathbb{N}^\star$,
	\item[--] $x_{(t=1)}$ is the known initial state,
	\item[--] $\Omega=\{x_1,\ldots,x_N\}$ is the finite set of states,
	\item[--] $\Gamma:\Omega\times\mathbb{R}^{m_{(t)}}\to \Omega$, $m_{(t)}=|U_{(t)}|\in\mathbb{N}^\star$, is a state-transition function mapping a state $x_{(t)} \in \Omega$ and an input vector $\vec{u}_{(t)}\in\mathbb{R}^{m_{(t)}}$ to a next state $x_{(t+1)}\in \Omega$. Each element of $\vec{u}_{(t)}$ qualifies the observation of an event supposed to act on the state $x_{(t)}$ to yield $x_{(t+1)}$. In this context, $\sqcup_{(x_{(t)}| x_{(t-1)})}=\{\vec{u}_1,\ldots, \vec{u}_n\}$ (denoted $\sqcup_{(t)}$ in the sequel) represents the set of input vectors whose values are supposed to trigger a state-transition from the state $x_{(t)}$ to the state $x_{(t+1)}$.
	\item[--] $G:\Omega\leadsto\psi_{\Omega}$, is a set-valued output function mapping each state $x_{(t)}$ to a set $\psi_{x_{(t)}}=\{\vec{y}_1,\ldots,\vec{y}_j\}$ of expected observations while being in state $x_{(t)}$. The $p$ elements of $\vec{y}_i \in \mathbb{R}^{p_{(t)}}$, $p_{(t)}=|Y_{(t)}|$, $1\leq i\leq j$, qualifies an expected physical effect while being in state $x_{(t)}$. 
	\item[--] $\mathcal{U}:\Omega\times\Omega\leadsto U$ is a function mapping a state-transition $(x_{(t-1)},x_{(t)})$ to the set of inputs $U_{(t)}$ needed to qualify this state-transition,
	\item[--] $\mathcal{Y}:\Omega\leadsto Y$ is a function mapping a state $x_{(t)}$ to the set of outputs $Y_{(t)}$ needed to qualify this state.
\end{itemize}

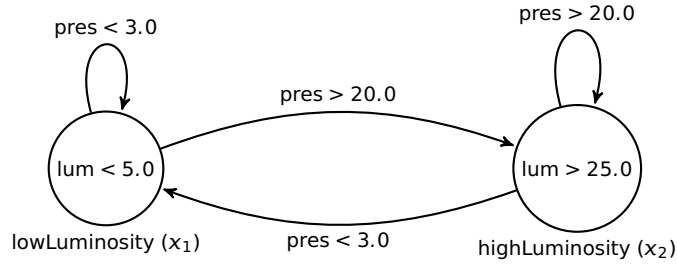
\begin{figure}[!ht]
	\centering
	\begin{tikzpicture}[->,>=stealth',shorten >=1pt,node distance=6.2cm,on grid,auto,thick,scale=0.75, every node/.style={scale=0.75}]
		\node[state, label=below:\low ($x_1$)] (LL) {$\l < 5.0$};
		\node[state, label=below:\high ($x_2$)] (HL) [right=of LL] {$\l > 25.0$};
		\path[->] (LL) edge [loop above] node {$\p < 3.0$} (LL);
		\path[->] (HL) edge [loop above] node {$\p > 20.0$} (HL);
		\path[->, bend left=20] (LL) edge node {$\p > 20.0$} (HL);
		\path[->, bend left=20] (HL) edge node {$\p < 3.0$} (LL);	
	\end{tikzpicture}	
	\caption{\label{_dfsa_bis}Deterministic model describing the expected evolution of a physical system (here a room whose luminosity level is supposed to change depending on whether an inhabitant is present or not).}	
\end{figure}

\noindent For instance, Fig.\ref{_dfsa_bis} depicts the expected behavior of a simple CPS whose purpose is to adjust the luminosity (physical property) of a room (the physical system) according to whether an inhabitant is present or not. While no inhabitant is present in the room (characterized by $pres<3$) then the value provided from the luminosity sensor should be less than 5 (characterized by $lum<5$). Here $m=p=1$.
\noindent Here, one may see a parallel with unit tests performed in software engineering for validating an algorithm. Some inputs are provided to the algorithm. The output, resulting from the treatment of these inputs by the algorithm, is compared with an expected output value. In this context, let us imagine one want to test that the sequence $(\vec{u}_t, \vec{y}_t)_{t=1}^T$ leads the sequence of states $(x_t)_{t=1}^T$ : the algorithm under test is the one controlling the CPS considered. The deterministic model of the expected behavior, here, plays the role of an unit test defined as follows :
\begin{flalign*}\nonumber
	\mathds{1}_{x_{(t)}}(x_{(t-1)},\vec{u}_{(t-1)})=&
	\left\{
		\begin{array}{l l}
			1 & \text{if } \vec{u}_{(t-1)}\in\sqcup_{(x_{(t)}|x_{(t-1)})}\\
			0 & \text{otherwise}	
		\end{array}
	\right.\\
	\mathds{1}_{x_{(t)}}(\vec{y}_{(t)})=&
	\left\{
		\begin{array}{l l}
			1 & \text{if } \vec{y}_{(t)}\in\psi_{x_{(t)}}\\
			0 & \text{otherwise}	
		\end{array}
	\right.
\end{flalign*}
\noindent The result of the test is then computed by :
\begin{equation}
\label{testeq}
\boxed{
 Test(\underbrace{(x_t)_{t=1}^T}_{\text{Expected result}}, \underbrace{(\vec{u}_t, \vec{y}_t)_{t=1}^T}_{Inputs}) = \prod_{t=2}^T\left( \mathds{1}_{x_{(t)}}(x_{(t-1)},\vec{u}_{(t-1)})\times \mathds{1}_{x_{(t)}}(\vec{y}_{(t)})\right)
 }
\end{equation}

\noindent \textbf{without room for tolerance towards uncertainties}, the test result can only be PASS or FAIL, i.e. $\in \{0,1\}$.\\

\noindent However, without being perfect, the luminosity level at 22.8 (state $x_2$ in Fig.\ref{_dfsa_bis}) when an inhabitant is present may be still acceptable and effectiveness $\in [0,1]$. So, one needs to extend the deterministic model allowing to define tolerances pertaining the following uncertainties:
\begin{itemize}
	\item [--] The \textit{aleatory uncertainties} which are most likely objective and relative to the natural variability of the physical properties of interest whose values over time are most likely distributed around an average value,
	\item [--] The \textit{epistemic uncertainties} which are most likely subjective and relative to users' satisfaction towards the physical system evolution.
\end{itemize}
\noindent Besides these uncertainties, one may also consider \textit{reliability uncertainties} such as:
\begin{itemize}
		\item [--] The \textit{spatial uncertainties} relative to the sensors location with respect to the physical property of interest,
		\item [--] The \textit{hardware uncertainties} relative to the sensors accuracy and resolution,
		\item [--] The \textit{model uncertainties} relative to the designer of the model and its expertise on the application domain.
\end{itemize}    

\subsection{Towards its formalization into the measure theory}
\label{sec:formal}
\noindent So as to handle the uncertainties previously described, we propose to generalize the deterministic model in the framework of the measure theory. Doing so, one can leverage the set of measures (probabilities, possibilities, beliefs, etc.) as a means of defining zones of viability from which one can reason in order to obtain the degree of effectiveness.

\subsubsection{Background}
\noindent Before formally generalizing the deterministic model in the framework of the measure theory and defining the degree of effectiveness, let us first review some key concepts of the measure theory. The reader is referred to the literature for details on this theory \cite{halmos2013measure}.
\begin{definition}\label{measure}
\noindent(\textbf{Measure}) Let $(\mathsf{X},\Sigma_{\mathsf{X}})$ be a measurable space where $\mathsf{X}$ is a countable set and $\Sigma_{\mathsf{X}}$ is a $\sigma$-algebra over $\mathsf{X}$. A function $\mu:\Sigma_{\mathsf{X}}\to\mathbb{R}_{\geq 0}$ is \cite{gilboa1994additive}:
	\begin{enumerate}
		\item Monotone if $\forall A,B\in\Sigma_{\mathsf{X}}, A\subseteq B$ implies $\mu(A)\leq\mu(B)$,
		\item Normalized if $\mu(\Sigma_{\mathsf{X}})=1$,
		\item Non-negative if $\forall A\in\Sigma_\mathsf{X}$, $\mu(A)\geq 0$,		
		\item Additive if $\forall A,B\in\Sigma_{\mathsf{X}}$ where $(A\cap B)=\emptyset$ then $\mu(A\cup B)=\mu(A)+\mu(B)$,
	\end{enumerate}
\end{definition} 

\noindent The function $\mu$ is said to be an \textit{additive measure} if it is monotone, non-negative, additive and $\mu(\emptyset)=0$. It is said to be a \textit{non-additive measure} if it is monotone, non-negative, non-additive and  $\mu(\emptyset)=0$. A measure is said to be a sub-measure if $\forall A,B\in\Sigma_{\mathsf{X}}$, $\mu(A\cup B)\leq\mu(A)+\mu(B)$.
\begin{definition}\label{measfunc}
	\noindent(\textbf{Measurable Function}) Let $(\mathsf{X},\Sigma_X)$ and $(\mathsf{Y},\Sigma_Y)$ be measurable spaces where $\mathsf{X}$ and  $\mathsf{Y}$ are countable sets and where $\Sigma_X$ and $\Sigma_Y$ are finite $\sigma$-algebras. A function $f:(\mathsf{X},\Sigma_X)\to(\mathsf{Y},\Sigma_Y)$ is said measurable if $f^{-1}(A)\in\Sigma_X$ $\forall A \in \Sigma_Y$.\\
\noindent For instance, let $\mathsf{X}=\{a,b,c,d\}$, and $\Sigma_X=\{\{a,b\},\{c,d\},\mathsf{X}, \emptyset\}$. Let $\mathsf{Y}=\{1,2,3\}$, and $\Sigma_Y=\{\{1\},\{2,3\},\{2\},\{1,3\},\{3\},\{1,2\},\mathsf{Y},\emptyset\}$. The function $f$, defined by $f(a)=1$, $f(b)=1$, $f(c)=2$ and $f(d)=2$, is measurable. Indeed, $f^{-1}(\{1\})=\{a,b\}\in\Sigma_X$, $f^{-1}(\{2,3\})=\{c,d\}\in\Sigma_X$, $f^{-1}(\{3\})=\emptyset\in\Sigma_X$, $f^{-1}(\{1,2\})=\mathsf{X}\in\Sigma_X$, etc.   
\end{definition}

\begin{definition}\label{ker}
	\noindent(\textbf{Kernel}) Let $(\mathsf{X},\Sigma_X)$ and $(\mathsf{Y},\Sigma_Y)$ be measurable spaces where $\mathsf{X}$ and  $\mathsf{Y}$ are countable sets and where $\Sigma_X$ and $\Sigma_Y$ are finite $\sigma$-algebras. A \textit{finite kernel} from $\mathsf{X}$ to $\mathsf{Y}$ is a function $K:\mathsf{X}\times\Sigma_Y\to\mathbb{R}_{\geq 0}$ that satisfies:
	\begin{itemize}
		\item[--] $\forall x \in\mathsf{X}, K(x,Y)$ is a measure on $(\mathsf{Y},\Sigma_Y)$,
		\item[--] $\forall Y \in\Sigma_Y, x\mapsto K(x,Y)$ is \textit{measurable}.
	\end{itemize}
\end{definition}
\noindent $\mathsf{X}$ and $\mathsf{Y}$ being countable sets, the kernel can be specified as a matrix $\{k(x,y):(x,y)\in\mathsf{X}\times\mathsf{Y}\}$. One can think of $k(x,y)$ as providing the conditional measure of $y$ given $x$. The kernel is referred to as a \textit{stochastic kernel} (a.k.a. Markov kernel or probability kernel) when $K:\mathsf{X}\times\Sigma_Y\to [0,1]$ and $\sum_{y\in\mathsf{Y}}k(x,y)=1, \forall x\in\mathsf{X}$, i.e. $K(x,\mathsf{Y})=1$ $\forall x\in\mathsf{X}$.  
\begin{definition}\label{kerprod}
\noindent(\textbf{Kernel Product}) Let $(\mathsf{X},\Sigma_X)$, $(\mathsf{Y},\Sigma_Y)$ and $(\mathsf{Z},\Sigma_Z)$ be measurable spaces. Let $k_1:\mathsf{X}\times\Sigma_Y\to\mathbb{R}_+$ and $k_2:(\mathsf{X}\times\mathsf{Y})\times\Sigma_Z\to\mathbb{R}_+$.
\end{definition}

\noindent Then, one can define the kernel product $k_1\otimes k_2:\mathsf{X}\times(\Sigma_Y\otimes\Sigma_Z)\to\mathbb{R}_+$ as a function of $k_1$ and $k_2$ \cite{klenke2013probability} where $\otimes$ is a product operator\footnote{In the literature, this operator is also known as fusion operator \cite{dubois2016basic}}.
\begin{theorem}\label{ionescu}
	\noindent(\textbf{Ionescu-Tulcea Extension Theorem}) \cite{shalizi2007almost} Let us consider a sequence of measurable spaces $(\mathsf{X}_n,\Sigma_n)_{n\in\mathbb{N}^\star}$. Let assume that for each $n$, there exists a kernel $K_n$ from $\times_{k=1}^{k=n-1}X_k$ to $X_n$. Then, for every sequence $(S_n)_{n\in\mathbb{N}^\star}$ taking values in $(X_n,\Sigma_n)$ there exists a unique measure $\mu(S_1,S_2,\ldots,S_n)=\otimes_{k=1}^{n}K_k$.   
\end{theorem}

\noindent With these key concepts defined, one can generalize the deterministic model described by Eq.\ref{_dfsa} in the measure theory framework.

\subsubsection{Generalizing the function \texorpdfstring{$\Gamma$}{} to the finite kernel \texorpdfstring{$K_S$}{}}
\noindent Let us consider the measurable spaces $(\Omega\times\mathbb{U},\Sigma_{\Omega\mathbb{U}})$ and $(\Omega,\Sigma_{\Omega})$ where $\Omega$ is the finite set of states, $\mathbb{U}\subseteq\mathbb{R}^m$ is the input vector, $\Sigma_{\Omega\mathbb{U}}$ is a finite $\sigma$-algebra on $\Omega\times\mathbb{U}$ and $\Sigma_{\Omega}$ is a finite $\sigma$-algebra on $\Omega$. A finite kernel $K_S$ from $\Omega\times\mathbb{U}$ to $\Omega$ is defined by (Definition.\ref{ker}):
\begin{equation}
	K_S:(\Omega\times\mathbb{U})\times\Sigma_\Omega\to [0,1]
\end{equation} 
\noindent $\Omega$ being a countable set, the kernel $K_S$ can be specified as a matrix $\{k_S((x,\vec{u}),x'):((x,\vec{u}),x')\in (\Omega\times\mathbb{U})\times\Omega\}$. \textbf{Think of $k_S((x,\vec{u}),A)$ as the conditional measure that the process will be in the state $A\subset\Omega$ at time $t$ given its state at time $t-1$ is $x\in\Omega$ and the input vector is $\vec{u}$}. Here, it is assumed that the state at time $t$ depends on the state at time $t-1$ and not on the previous states $t-2,t-3,\ldots,t_1$ (first order Markov property). Thus, ${\{\mathfrak{X}_t,\mathfrak{U}_t\}}_{t\in\mathbb{N}^\star}$, where $\mathfrak{X}_t$ and $\mathfrak{U}_t$ are random variables taking values in $\Omega$ and $\mathbb{U}$ respectively, is a \textit{chain} with kernels ${K_S}_{(t\geq 2)}$ and initial distribution $\pi$ where $\pi:\Sigma_\Omega\to [0,1]$ is a measure on $(\Omega,\Sigma_\Omega)$ at $t=1$.

\subsubsection{Generalizing the function \texorpdfstring{$G$}{} to the finite kernel \texorpdfstring{$K_E$}{}}
\noindent Let us consider the measurable spaces $(\Omega,\Sigma_\Omega)$ and $(\mathbb{Y},\Sigma_\mathbb{Y})$ where $\mathbb{Y}\subseteq\mathbb{R}^p$ is the output vector and $\Sigma_\mathbb{Y}$ is a $\sigma$-algeba on $\mathsf{\mathbb{Y}}$. A finite kernel $K_E$ from $(\Omega,\Sigma_\Omega)$ to $(\mathbb{Y},\Sigma_\mathbb{Y})$ is defined by:	
\begin{equation}
K_E:\Omega\times\Sigma_\mathbb{Y}\to [0,1]
\end{equation} 	
\noindent \textbf{Think of ${K_E}_{(t)}(x,\vec{y})$ as the conditional measure that the process is in the state $x$ at time $t$ given the output vector $\vec{y}\subset\mathbb{Y}$ at time $t$}.
\begin{figure}[!ht]
\centering
\begin{tikzpicture}[thick,scale=0.75, every node/.style={scale=0.75}]
	\node at (-4.75,1.9) {$x_{(t-1)}$};
	\node at (-0.5,1.9) {$x_{(t)}$};
	\node at (-4.75,4) {$\vec{u}_{(t-1)}$};
	\node at (-0.5,-0.25) {$\vec{y}_{(t)}$};
	\node at (-2.3,-1.55) {\Huge$\otimes$};
	\node at (-4.8,-1.55) {\color{blue}$k_S\left((x_{(t-1)},\vec{u}_{(t-1)}),x_{(t)}\right)$};
	\node at (-4.8,-2.1) {State prediction};
	\node at (-0.45,-2.1) {State emission};
	\node at (-0.5,-1.55) {\color{red}$k_E(x_{(t)},\vec{y}_{(t)})$};
	\draw  [line width = 0.5mm](-4.75,1.9) ellipse (0.8 and 0.8);
	\draw  [line width = 0.5mm](-4.75,4) ellipse (0.8 and 0.8);
	\draw  [line width = 0.5mm](-0.5,-0.25) ellipse (0.8 and 0.8);
	\draw  [line width = 0.5mm](-0.5,1.9) ellipse (0.8 and 0.8);
	\draw [->, line width = 0.4mm](-3.95,1.9) -- (-1.3,1.9);
	\draw [->, line width = 0.4mm](-3.95,4) -- (-1.3,2.1);
	\draw [->, line width = 0.5mm](-0.5,1.1) -- (-0.5,0.55);
		
	\draw  [black!20,line width = 0.5mm](-8.9,1.85) ellipse (0.8 and 0.8);
	\draw  [black!20,line width = 0.5mm](-8.9,4) ellipse (0.8 and 0.8);
	\draw  [black!20,line width = 0.5mm](3.6,1.9) ellipse (0.8 and 0.8);
	\draw  [black!20,line width = 0.5mm](-0.5,4) ellipse (0.8 and .8);	
	\draw [->,black!20](-8.05,1.9) -- (-5.55,1.9);
	\draw [->,black!20](0.3,4) -- (2.8,2.15);
	\draw [->,black!20](0.3,1.9) -- (2.8,1.9);	
	\draw  [black!20,line width = 0.5mm](-4.75,-0.25) ellipse (0.8 and .8);
	\draw  [black!20,line width = 0.5mm](-8.9,-0.25) ellipse (0.8 and .8);
	\draw  [black!20,line width = 0.5mm](3.6,-0.25) ellipse (0.8 and .8);
	\draw  [black!20,line width = 0.5mm](3.6,4) ellipse (0.8 and .8);
	\draw  [->,black!20](3.6,1.1) -- (3.6,0.55);
	\draw  [->,black!20](-4.75,1.1) -- (-4.75,0.55);
	\draw  [->,black!20](-8.9,1.1) -- (-8.9,0.55);	
	\node at (-8.9,1.85) {\color{black!20}$x_{(t-2)}$};
	\node at (-10.2,1.85) {\color{black!20}$\cdots$};
	\node at (-8.9,-0.25) {\color{black!20}$\vec{y}_{(t-2)}$};
	\node at (-4.75,-0.25){\color{black!20}$\vec{y}_{(t-1)}$};
	\node at (-8.9,4) {\color{black!20}$u_{(t-2)}$};
	\node at (-0.5,4) {\color{black!20}$u_{(t)}$};
	\node at (3.6,4) {\color{black!20}$u_{(t+1)}$};
	\node at (3.6,1.85) {\color{black!20}$x_{(t+1)}$};
	\node at (4.9,1.85) {\color{black!20}$\cdots$};
	\node at (3.6,-0.25) {\color{black!20}$\vec{y}_{(t+1)}$};
	\draw [->,gray!20](-8.05,4) -- (-5.55,2.1);
	\draw  [red,line width=0.3mm,rounded corners=3mm](-1.75,3.15) rectangle (0.5,-1.2);	
	\draw [blue, line width=0.3mm,rounded corners=3mm](-6,3.5) node (v1) {} -- (-6,1) -- (0.65,1) -- (0.65,3) -- (-3.5,3) -- (-3.5,5) -- (-6,5) -- (-6,3.5);
\end{tikzpicture}
\caption{\label{kernels} Markov transition kernel $K_S\otimes K_E$.}
\end{figure}
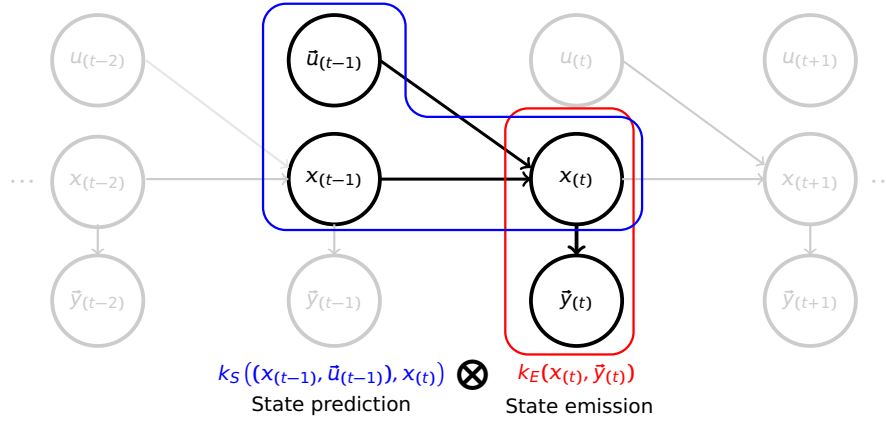
\noindent Per Definition.\ref{kerprod}, at each time $t$, the \textit{Markov transition kernel} $K_S\otimes K_E:(\Omega\times\mathbb{U})\times(\Sigma_\Omega\otimes\Sigma_\mathbb{Y})\to [0,1]$, is a function of $K_S$ and $K_E$ (Fig.\ref{kernels}). Think of $K_S\otimes K_E$ as the conditional measure of $x_{(t)}$ given $\vec{y}_{(t)}$, $\vec{u}_{(t-1)}$ and $x_{(t-1)}$.

\subsection{Transferable Belief Model (TBM)}
\subsubsection{Basic definitions and notations}
\noindent Let us consider $\Omega=\{x_1,x_2,\ldots,x_N\}$ the discrete \textit{frame of discernment} (FoD) representing the states of a physical system where $x_1, x_2,\ldots$ are \textit{hypothesis}. In this paper, hypothesis are supposed to be exhaustive and exclusive, i.e. the system cannot be in two states at once. A \textit{mass function} $m^\Omega$, a.k.a. Basic Belief Assignment (BBA) is defined by:
\begin{equation}
m^\Omega:2^\Omega\rightarrow [0,1] 
\end{equation}
\noindent where $2^\Omega=\{\emptyset,\{x_1\},\{x_2\},\{x_1,x_2\},\{x_3\},\{x_1,x_3\},\{x_2,x_3\},\{x_1,x_2,x_3\},...,\Omega\}$. A BBA is \textit{non-additive}, i.e \begin{equation}m^\Omega(\{x_1,x_2\})\neq m^\Omega(\{x_1\})+m^\Omega(\{x_2\})\end{equation} This is a fundamental difference with probability theory. A proposition $A=\{x_1,x_2\}\in 2^\Omega$ explicitly represents the doubt between hypothesis composing $A$ and the mass of belief $m^\Omega(A)$ assigned to $A$ is not informative regarding the elements of $A$.\\
\noindent A BBA is a set of belief masses concerning propositions $A\in 2^\Omega$ verifying:
\begin{equation}
\sum_{A\in 2^\Omega}m^\Omega(A)=1
\end{equation} 
\noindent $A\in 2^\Omega$ is a \textit{focal element} of the BBA if $m^\Omega(A)>0$.
\medskip

\noindent\fbox{\begin{minipage}{0.972\linewidth}		
\noindent \textbf{In the Dempster-Shafer theory of evidence $m^\Omega(\emptyset)$ is constrained to 0. This constraint is relaxed in TBM \cite{smets1994transferable} where $m^\Omega(\emptyset)>0$ is given different interpretations \cite{lefevre2002belief}:}
\begin{enumerate}
	\item Inaccuracy of the sensors measurements (Observations),
	\item Incompleteness of the model leading to non-exhaustive FoD.
\end{enumerate}
\end{minipage}}
\subsubsection{Belief functions}
\noindent BBAs can be transformed to one-to-one relationships \cite{smets1994transferable} representing the same information (a.k.a. belief functions), albeit in different forms. Some are described hereafter.    
\begin{itemize}
	\item[--] \textbf{Plausibility $pl$} where \begin{equation}pl^\Omega(A)=\sum_{B\cap A\neq 0}m^\Omega(B), \forall A,B\in 2^\Omega\end{equation}	
	\noindent and reversely \begin{equation}\label{pm}m^{\Omega}(A)=\sum_{B\subseteq A}(-1)^{|A|-|B|+1}pl^{\Omega}(\bar{B}), \forall A,B\in 2^\Omega\end{equation}	
	\item[--] \textbf{Belief $bel$} where  \begin{equation}bel^\Omega(A)=\sum_{\emptyset\neq B\subseteq A}m^\Omega(B), \forall A,B\in 2^\Omega \text{, with $m^\Omega$ normal, i.e. $m^\Omega(\emptyset)=0$.} \end{equation}
	\item[--] \textbf{Commonality $q$} where \begin{equation}\label{q}q^\Omega(A)=\sum_{B\supseteq A}m^\Omega(B), \forall A,B\in 2^\Omega\end{equation}
	\noindent and reversely \begin{equation}\label{qm}m^{\Omega}(A)=\sum_{A\subseteq B}(-1)^{|B|-|A|}q^{\Omega}(B), \forall A,B\in 2^\Omega\end{equation}
\end{itemize}
\subsubsection{CRC/DRC combination rules}
\noindent There have been many combination rules proposed in the literature \cite{smets2007analyzing}. In the sequel, we do consider the Conjunctive Rule of Combination (CRC) and the Disjunctive Rule of Combination (DRC).
\medskip

\begin{definition}\textbf{Conjunctive Rule of Combination (CRC). }
\noindent Let us consider two BBAs defined by $m^\Omega_1$ and $m^\Omega_2$.\\
\noindent\fbox{\begin{minipage}{0.972\linewidth}	
\noindent \textbf{Assuming their sources are independent and \textit{reliable} then the unnormalized \textit{conjunctive rule of combination} (CRC $\crc$) can be used as follows \cite{smets2008belief}:}
\end{minipage}}

\medskip 
$\forall A\in 2^\Omega$, by:
\begin{equation}
m^\Omega_{\footnotesize 1\crc 2}(A) = \sum_{B\cap C=A} m^\Omega_1(B).m^\Omega_2(C) \text{, } A\in 2^\Omega 
\end{equation}
\begin{equation}\label{_crc}
q^\Omega_{\footnotesize 1\crc 2}(A) = q^\Omega_1(A).q^\Omega_2(A)
\end{equation}
\end{definition}
\medskip

\noindent This combination may result in a \textit{sub-normal} BBA, i.e. $m^\Omega(\emptyset)>0$. The mass of conflict is given by:
\begin{equation}
m^\Omega_{\footnotesize 1\crc 2}(\emptyset)=\sum_{A\cap B=\emptyset}m^\Omega_1(A).m^\Omega_2(B) \text{, } A,B\in 2^\Omega 
\end{equation}

\noindent It is worth noting that the CRC can be computed from commonality functions:
\begin{equation}\label{conflict}
m^\Omega_{\footnotesize 1\crc 2}(\emptyset)=1+\sum_{A\in 2^\Omega, A\neq\emptyset} (-1)^{|A|}\cdot q^\Omega_1(A).q^\Omega_2(A)
\end{equation}
\medskip

\begin{definition}\textbf{Disjunctive Rule of Combination (DRC). }
\label{__crc}
\noindent Let us consider two BBAs defined by $m^\Omega_1$ and $m^\Omega_2$.\\
\noindent\fbox{\begin{minipage}{0.972\linewidth}	
\noindent \textbf{Assuming their sources are independent and at least one source is reliable, then the unnormalized \textit{disjunctive rule of combination} (DRC $\drc$) can be used as follows \cite{smets2008belief}:}
\end{minipage}}
\medskip 

\begin{equation}
m^\Omega_{\footnotesize 1\drc 2}(A) = \sum_{B\cup C=A} m^\Omega_1(B).m^\Omega_2(C) \text{, } A\in 2^\Omega 
\end{equation}
\begin{equation}\label{_drc}
bel^\Omega_{\footnotesize 1\drc 2}(A) = bel^\Omega_1(A).bel^\Omega_2(A)
\end{equation}
\end{definition}

\section{Degree of effectiveness}
\subsection{Formalization in the measure theory}
\noindent On the basis of the formalization of the deterministic model of the CPS expected behavior in the measure theory described in \ref{sec:formal} and extending Eq.\ref{testeq}, the degree of effectiveness can be formulated as follows:
\begin{definition}\label{def}The degree of effectiveness is a function $\delta:\Omega_{(t)}\times {(\mathbb{U}\times\mathbb{Y})}_{(t)} \to [0,1], t\in\mathbb{N}^\star$ such that given the state sequence $(x_t)_{t=1}^T$ and the observation sequence $(\vec{u}_t,\vec{y}_t)_{t=1}^T$, the degree of effectiveness $\delta\left((x_t)_{t=1}^T,(\vec{u}_t,\vec{y}_t)_{t=1}^T\right)$ is given by:
\begin{equation}\label{doe}
	\boxed{
		\small
		\delta\left((x_t)_{t=1}^T,(\vec{u}_t,\vec{y}_t)_{t=1}^T\right)=\underbrace{\vphantom{\bigotimes_{t=2}^T K_S}\pi(x_1)\otimes K_E(x_1,\vec{y}_1)}_{\text{Initialization}}\otimes\left[\underbrace{\bigotimes_{t=2}^T K_S\left((x_{(t-1)},\vec{u}_{(t-1)}),x_{(t)}\right)\otimes K_E\left(x_{(t)},\vec{y}_{(t)}\right)}_{\text{prediction - update mechanism}}\right]
	}
\end{equation}
\end{definition}

\noindent The process consists in propagating the measure over the state sequence $(x_t)_{t=1}^T$. At each time $t$, it satisfies a "prediction ($K_S$) - update ($K_E$)" mechanism. It can be understood as the 'likelihood' of the state sequence $(x_t)_{t=1}^T$ to have been produced by the observation sequence $(\vec{u}_t,\vec{y}_t)_{t=1}^T$. $\pi(x_i)$ gives the measure of the state sequence to start by the state $x_i$.\\
\medskip

\noindent As Eq.\ref{doe} provides the degree of effectiveness for one possible sequence of states, one needs to find the sequence of states leading the highest degree of effectiveness over all the possible sequences of states given the observation sequence, i.e. 

\begin{equation}\label{_doe}
	\boxed{
		\small
		\delta\left((\vec{u}_t,\vec{y}_t)_{t=1}^T\right)= \max_{(x_t)_{t=1}^T}\delta\left((x_t)_{t=1}^T,(\vec{u}_t,\vec{y}_t)_{t=1}^T\right)
	}
\end{equation}
\medskip

\begin{note}
\noindent The chain ${\{\mathfrak{X}_t,\mathfrak{U}_t,\mathfrak{Y}_t\}}_{t\in\mathbb{N}^\star}$ where $\mathfrak{X}_t$,$\mathfrak{U}_t$ and $\mathfrak{Y}_t$ are random variables taking values in $\Omega$, $\mathbb{R}^{m_{(t)}}$ and $\mathbb{R}^{p_{(t)}}$ respectively, with transition kernel $K_S\otimes K_E$ and initial distribution $\pi\otimes K_E$, is an \textit{Input/Output Hidden Markov Model} (IOHMM)\footnote{In this paper we do assume that the state at time $t$ only depends on the state at time $t-1$ and not on the previous states $t-1,t-2,\ldots,t_1$ (first order Markov chain).} \cite{Frasconi1997} (claim derived from \cite{cappe2009inference}).
\end{note}
\medskip
\noindent Following this definition, an observation sequence $(\vec{u}_t,\vec{y}_t)_{t=1}^T$ is said \textit{perfect} when
\begin{equation}\nonumber
\delta\left((\vec{u}_t,\vec{y}_t)_{t=1}^T\right)=1 \text{ ,} \forall T\in\mathbb{N}^\star
\end{equation}

\noindent To be more precise about the meaning we seek to give to the degree of effectiveness, we borrow some terminology employed in the viability theory \cite{aubin:inria-00636570}. Let us consider that the constraints the physical system evolution has to comply with are encoded into information $I$. Then, the following definitions are adopted: 

\begin{definition}\noindent\label{comfort}A \textbf{zone of comfort} $\cf_{(t)}(I)$ associated to an event $E\in\Sigma_\Omega$ at time $t$ corresponds to the set $\mathcal{C}\subset\mathbb{U}\times\mathbb{Y}$ of values for which \textbf{the event $E$ is certain according to I}, such that:
\begin{equation}\delta\left((x_t)_{t=1}^T,(\vec{u}_t,\vec{y}_t)_{t=1}^T\right)=1\end{equation}
\end{definition}

\begin{definition}\noindent A \textbf{zone of tolerance} $\u_{(t)}(I)$ associated to an event $E\in\Sigma_\Omega$ at time $t$ corresponds to the set $\mathcal{T}\subset\mathbb{U}\times\mathbb{Y}$ of values for which \textbf{the event $E$ is uncertain according to I}, such that:
\begin{equation}0<\delta\left((x_t)_{t=1}^T,(\vec{u}_t,\vec{y}_t)_{t=1}^T\right)< 1\end{equation}
\end{definition}	

\begin{definition}\noindent A \textbf{zone of viability} $\v_{(t)}(I)$ associated to an event $E\in\Sigma_\Omega$ at time $t$ corresponds to the union of $\cf_{(t)}(I)$ and $\u_{(t)}(I)$. Note that outside the zone of viability, \textbf{the event $E$ is impossible according to I}:
\begin{equation}\delta\left((x_t)_{t=1}^T,(\vec{u}_t,\vec{y}_t)_{t=1}^T\right)=0\end{equation}	 \end{definition}

\noindent\fbox{\begin{minipage}{0.972\linewidth}		
	\textbf{Thus, the degree of effectiveness determines zones of viability according to the model, i.e. it determines the \textit{boundaries} of the states defined in the model. When $\delta(.)=0$, one faces a model breakdown, i.e. the state of the system is outside the boundaries of the states defined in the model}.
\end{minipage}}
\medskip

\begin{figure}[!ht]
	\centering
	\includegraphics[scale=0.2]{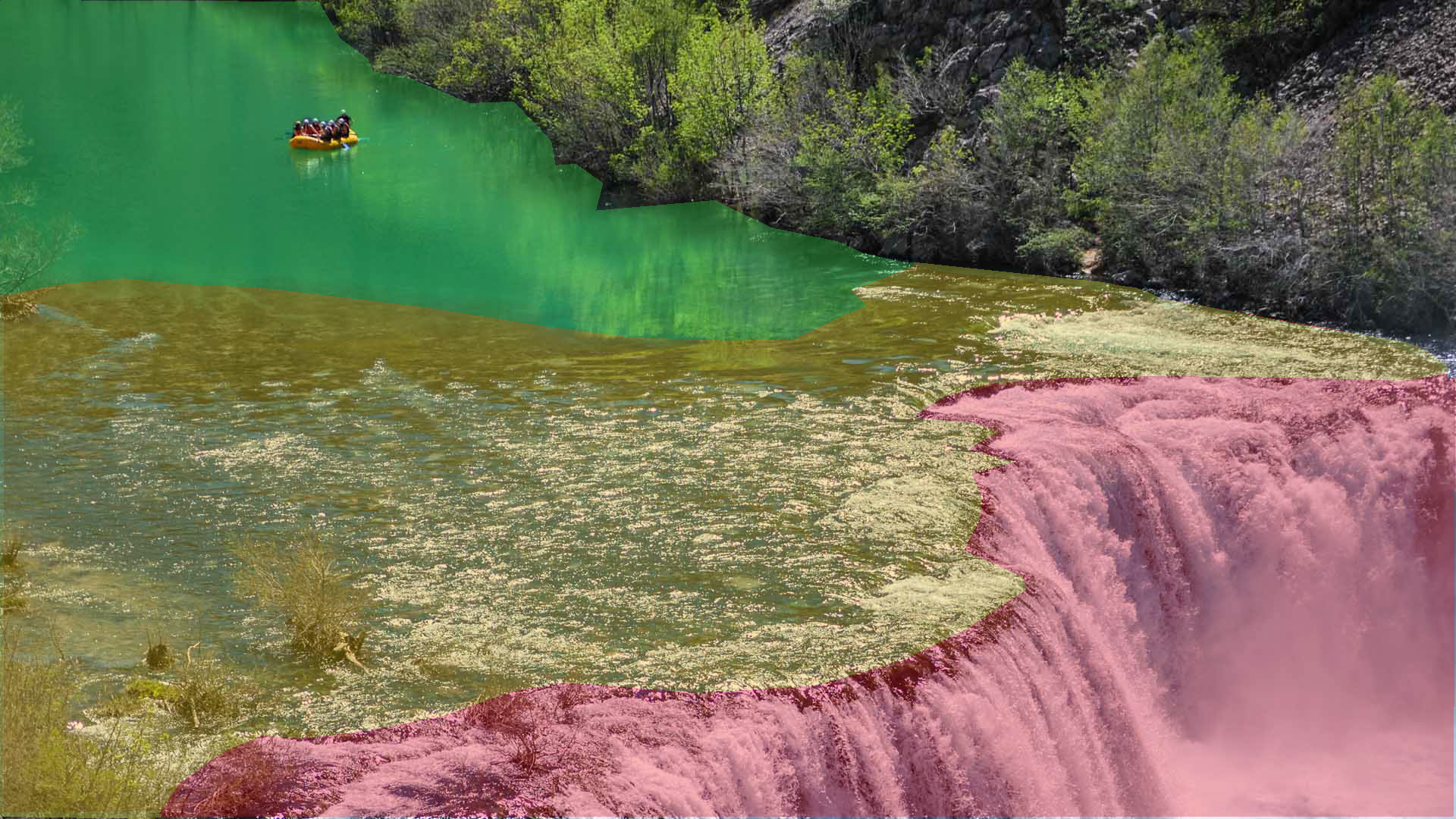}
	\put(-320,150){\textbf{Zone of comfort}}
	\put(-250,90){\textbf{Zone of tolerance}}
	\put(-30,20){\Huge{?}}
	\caption{\label{viabzone} Example of a viability zone where the event E can be stated as "\textit{passenger of the ship are safe}"\\Photo: Courtesy of Raftrek Travel \cite{raftrek}.}
\end{figure}

\noindent The Fig.\ref{viabzone} provides an illustrative example. Here, the event E can be stated as "\textit{the passengers of the ship are safe}". An input of the model might be the geographic position of the ship (latitude/longitude), while the output might be the heart rate of the passengers. Within the zone of comfort one can be certain that the passengers are safe, i.e. their heart rate is at the expected level. Within the zone of tolerance, passengers may suffer from disturbances and their safety is at risk, i.e. their heart rate is higher than expected. The ship is not supposed to go outside the boundary of the zone of viability\ldots

\newmdenv[linecolor=yellow!30,backgroundcolor=yellow!30]{st}
\newmdenv[linecolor=blue!10,backgroundcolor=blue!10]{matpi}
\newmdenv[linecolor=green!10,backgroundcolor=green!10]{mata}
\newmdenv[linecolor=orange!10,backgroundcolor=orange!10]{matb}

\subsection{Application to the Transferable Belief Model}
\label{sec:eviohmm}
\noindent Per Eq.\ref{doe} and Eq.\ref{_doe}, by replacing kernels $K_S$ and $K_E$ with BBAs, the computation of the degree of effectiveness can be factored as follows : 

\begin{equation}\label{TBMEff}
\boxed{
  \begin{minipage}{\displaywidth}
	$\delta^{\Omega_T}[(\vec{u}_t,\vec{y}_t)_{t=1}^T] = \max_{(x_t)_{t=1}^T}$\\
	$\left(m^\Omega_\pi(\{x_1\}) \times m^{\Omega_{1}}[\vec{y}_{1}](\{x_1\})\times\left[\prod_{t=2}^T m^{\Omega_{(t)}|\Omega_{(t-1)}}[\{x_{(t-1)}\}, \vec{u}_{(t-1)}](\{x_{(t)}\})\times m^{\Omega_{(t)}}[\vec{y}_{(t)}](\{x_{(t)}\})\right]\right)$
  \end{minipage}	
}
\end{equation}
\noindent where $m^{\Omega_{(t)}|\Omega_{(t-1)}}[A](B)$ represents a belief function defined on $\Omega_{(t)}$ conditionally to the subset $A\subseteq\Omega_{(t-1)}$. For the sake of simplicity, $m^{\Omega_{(t)}|\Omega_{(t-1)}}[A](B)$ is replaced by $m^{\Omega_{(t)}}[A](B)$ in the sequel. It is worth noting that in Eq.\ref{TBMEff} the masses involved in the computation are supposed to be known.

\subsection{Evidential Input/Output Hidden Markov Model (Ev-IOHMM)}
\label{eviohmm}
\noindent The work presented in this paper extends works done on Evidential HMM (Ev-HMM) \cite{ramasso2017inference}\cite{ramasso2007forward} and probabilistic Input/Output HMM (IOHMM) \cite{bengio1995input}. In the sequel, we do assume the reader is familiar with basics in HMM.\\

\medskip

\noindent Formally, an Evidential IOHMM (Ev-IOHMM) is defined by the tuple $\lambda=<\Omega,A,\vec{B}, m^\Omega_\pi>$ where:
\begin{itemize}[leftmargin=0.3cm]
	\item[--]
	$\mathbm{\Omega}=\{x_1,x_2,\ldots,x_N\}$ is the finite set of hidden states, i.e. the frame of discernment,
	\item[--] 
    $\mathbm{\vec{B}}$ is the $2^{|\Omega|}$ emission vector whose elements represent the beliefs conditional to the output value $\vec{y}$. For instance, $m^{\Omega_{(t)}}[\vec{y}_{(t)}](\{x_1\}_{(t)})$ represents the belief in $\{x_1\}$ at time $t$ given the output observation $\vec{y}_{(t)}$ at time t. 

	\item[--] 
$\mathbm{A}$ is the $|\Omega|\times 2^{|\Omega|}$ state-transition matrix. There is one row per singleton $\in\Omega$. Each row of the matrix is a BBA whose elements represent the belief in transiting from the singleton to this element. For instance, in Ev-HMM, $m^{\Omega_{(t)}}[\{x_1\}_{(t-1)}](\{x_2\}_{(t)})$ represents the belief in transitioning to state $\{x_2\}$ at time $t$ given the state at time $t-1$ was $\{x_2\}$. Here the belief is only conditional to the previous state. \textbf{In Ev-IOHMM, the belief in transitioning from one state to another is also conditional to an input. For instance, $m^{\Omega_{(t)}}[\{x_1\}_{(t-1)}, \vec{u}_{(t-1)}](\{x_2\}_{(t)})$ represents the belief in transitioning to state $\{x_2\}$ at time $t$ given the state at time $t-1$ was $\{x_2\}$ and the input value was $\vec{u}$.}.

	\item[--] 
	$\mathbm{m^\Omega_\pi}$ is a \textit{vacuous} BBA, i.e. $m^\Omega_\pi(\Omega)=1$ meant to indicate that one has no information on the initial state of the system.
\end{itemize}

\begin{note}
In real life applications, BBAs are often not directly available. Only the probability or the possibility values computed from observations are available on the singletons. So, the model is extended with a vector $\vec{B}'$ and a matrix $A'$ whose elements describe probability density functions or distributions of possibility :

\begin{itemize}[leftmargin=0.3cm]
	\item[--] 
    $\mathbm{\vec{B}'}$ is a $|\Omega|$ vector where each element $b_i$, $(1\leq i\leq |\Omega|)$ is a probability density function or distribution of possibility. For instance, ${b_i}(\vec{y})=p(\vec{y}_{(t)} = \vec{y} | x_{(t)} = i)$ denotes the probability of observing the output vector $\vec{y}$ at time $t$ given the state is $x_i$ at time $t$. 

	\item[--] 
$\mathbm{A'}$ is a $|\Omega|\times |\Omega|$ matrix where each element $a_{ij}$, $(1\leq i,j\leq |\Omega|)$ is a probability density function or distribution of possibility. For instance, $a_{ij}(\vec{u})=p(x_{(t+1)} = j | x_{(t)}=i, \vec{u}_{(t)}=\vec{u})$ denotes the probability of transiting to $x_{(t+1)}$ at time $t+1$, given the state is $x_i$ at time $t$ and the input vector is $\vec{y}$ at time $t$.
\end{itemize}
\medskip

\noindent The distributions in $A'$ and $B'$ can be defined by the designer of the model when distributions represent, for instance, users' preferences or specific behavioral requirements/constraints. However, to date, no effort has been put on learning the model parameters from observations (following what has been done in \cite{ramasso2017inference} on the Ev-HMM or in \cite{bengio1996input} for the IO-HMM).
\end{note}
\medskip

\noindent\fbox{\begin{minipage}{0.972\linewidth}	
\noindent The HMM modeling framework and derivatives rely on computationally efficient reasoning algorithms \cite{stamp2015revealing}. Among these algorithms, \textbf{the forward algorithm offers a solution to the \textit{evaluation problem}. It computes the "likelihood" of the observation sequences $(\vec{u}_t)_{t=1}^T$ and $(\vec{y}_t)_{t=1}^T$ to have been produced by the model by taking into account all the possible underlying state sequences}. In other words, it provides a solution to the equation Eq.\ref{TBMEff}. 
\end{minipage}}

\subsubsection{State prediction}
\label{sp}

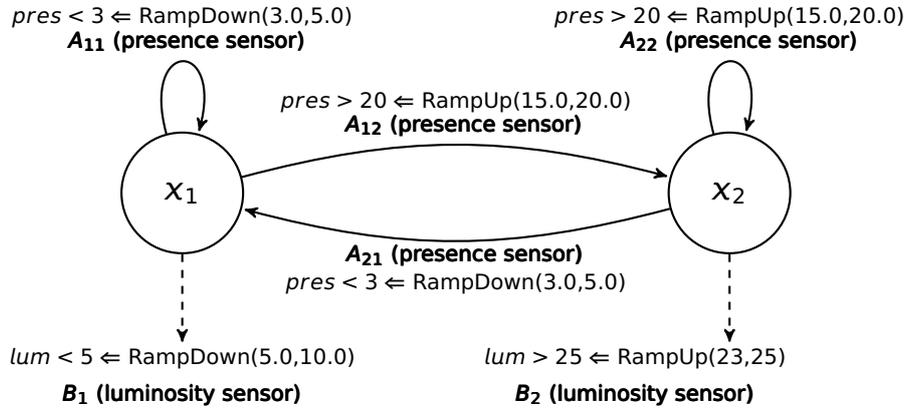
\begin{figure}[!ht]
\centering
\begin{tikzpicture}[->,>=stealth',shorten >=1pt,node distance=9.0cm, line width=1.pt, thick,scale=0.8, every node/.style={scale=0.8}]
\tikzstyle{every state}=[align=center]
\node[state,minimum size=2cm]	(LL)  {\Large $x_1$};
\node[state,minimum size=2cm]	(HL) [right of=LL]  	{\Large $x_2$};
\path [bend left=15, align=center] 	(LL)	edge	node[above=0.33]{$pres > 20 \Leftarrow$ RampUp(15.0,20.0)} (HL);
\path [bend left=15, align=center] 	(HL)	edge	node[below=0.32]{$pres < 3 \Leftarrow$ RampDown(3.0,5.0)} (LL);
\path (LL) 	edge  [loop above, align=center] 	node [above=0.35] {$pres < 3 \Leftarrow$ RampDown(3.0,5.0)} (LL);
\path (HL) 	edge  [loop above, align=center] 	node [above=0.35] {$pres > 20 \Leftarrow$ RampUp(15.0,20.0)} (HL);
\node[state,draw=none] (e1) [below =1.2cm of LL] {};
\node[state,draw=none] (e2) [below =1.2cm of HL] {};
\path(LL) edge [->, align=center, dashed] node {} (e1);
\path(HL) edge [->, align=center, dashed] node {} (e2);
\node [align=center, label=below:\pmb{$B_1$ (luminosity  sensor)}] at (0,-2.7375) {$lum < 5 \Leftarrow$ RampDown(5.0,10.0)};
\node [align=center, label=below:\pmb{$B_2$ (luminosity  sensor)}] at (7.4477,-2.7375) {$lum > 25 \Leftarrow$ RampUp(23,25)};
\node at (0.0818,2.5114) {\pmb{$A_{11}$ (presence sensor)}};
\node at (4.6352,1.1246) {\pmb{$A_{12}$ (presence sensor)}};
\node at (9.1676,2.5114) {\pmb{$A_{22}$ (presence sensor)}};
\node at (4.652,-1.0337) {\pmb{$A_{21}$ (presence sensor)}};
\end{tikzpicture}
	\caption{Model extending the model depicted in Fig.\ref{_dfsa_bis} with tolerances towards uncertainties described as distributions of possibility. This model can be read as follows: while a presence is detected in the room (presence sensor value $>$ 20.0), the value of the luminosity sensor must be higher than 25.0 (state $x_2$). Otherwise, if no presence is detected in the room (presence sensor value $<$ 3.0), the value of the luminosity sensor should be lower than 5.0 (state $x_1$). Uncertainties are handled through tolerances used to relax these constraints (see Fig.\ref{example102}).  }
	\label{example101}
\end{figure}

\begin{figure}[!ht]
	\centering
	\tikzset{every picture/.style={line width=1.0pt}}     
	\begin{tikzpicture}[x=0.75pt,y=0.75pt,yscale=-1,xscale=1]
	
\draw  (23,110.8) -- (295.5,110.8)(50.25,19) -- (50.25,121) (288.5,105.8) -- (295.5,110.8) -- (288.5,115.8) (45.25,26) -- (50.25,19) -- (55.25,26)  ;
\draw  (23,260.8) -- (295.5,260.8)(50.25,169) -- (50.25,271) (288.5,255.8) -- (295.5,260.8) -- (288.5,265.8) (45.25,176) -- (50.25,169) -- (55.25,176)  ;
\draw  (362,110.8) -- (634.5,110.8)(389.25,19) -- (389.25,121) (627.5,105.8) -- (634.5,110.8) -- (627.5,115.8) (384.25,26) -- (389.25,19) -- (394.25,26)  ;
\draw  (362,260.8) -- (634.5,260.8)(389.25,169) -- (389.25,271) (627.5,255.8) -- (634.5,260.8) -- (627.5,265.8) (384.25,176) -- (389.25,169) -- (394.25,176)  ;
\draw  [dash pattern={on 4.5pt off 4.5pt}]  (48.14,39.29) -- (290.14,39.29) ;
\draw  [dash pattern={on 4.5pt off 4.5pt}]  (48.14,75.29) -- (290.14,75.29) ;
\draw  [dash pattern={on 4.5pt off 4.5pt}]  (48.14,189.29) -- (290.14,189.29) ;
\draw  [dash pattern={on 4.5pt off 4.5pt}]  (48.14,225.29) -- (290.14,225.29) ;
\draw  [dash pattern={on 4.5pt off 4.5pt}]  (388.14,189.29) -- (630.14,189.29) ;
\draw  [dash pattern={on 4.5pt off 4.5pt}]  (388.14,225.29) -- (630.14,225.29) ;
\draw  [dash pattern={on 4.5pt off 4.5pt}]  (388.14,39.29) -- (630.14,39.29) ;
\draw  [dash pattern={on 4.5pt off 4.5pt}]  (388.14,75.29) -- (630.14,75.29) ;
\draw [color={rgb, 255:red, 208; green, 2; blue, 27 }  ,draw opacity=1 ]   (48.14,39.29) -- (140.14,39.29) ;
\draw [color={rgb, 255:red, 208; green, 2; blue, 27 }  ,draw opacity=1 ]   (140.14,39.29) -- (200.14,110.29) ;
\draw [color={rgb, 255:red, 208; green, 2; blue, 27 }  ,draw opacity=1 ]   (200.14,110.29) -- (288.14,110.29) ;
\draw [color={rgb, 255:red, 208; green, 2; blue, 27 }  ,draw opacity=1 ]   (230.14,189.29) -- (155.14,260.29) ;
\draw [color={rgb, 255:red, 208; green, 2; blue, 27 }  ,draw opacity=1 ]   (50.25,260.8) -- (155.14,260.29) ;
\draw [color={rgb, 255:red, 208; green, 2; blue, 27 }  ,draw opacity=1 ]   (290.14,189.29) -- (230.14,189.29) ;
\draw [color={rgb, 255:red, 208; green, 2; blue, 27 }  ,draw opacity=1 ]   (388.14,39.29) -- (465.14,39.57) ;
\draw [color={rgb, 255:red, 208; green, 2; blue, 27 }  ,draw opacity=1 ]   (465.14,39.57) -- (540.14,110.57) ;
\draw [color={rgb, 255:red, 208; green, 2; blue, 27 }  ,draw opacity=1 ]   (540.14,110.57) -- (627.14,110.57) ;
\draw [color={rgb, 255:red, 208; green, 2; blue, 27 }  ,draw opacity=1 ]   (510.14,260.57) -- (389.25,260.8) ;
\draw [color={rgb, 255:red, 208; green, 2; blue, 27 }  ,draw opacity=1 ]   (570.14,188.57) -- (510.14,260.57) ;
\draw [color={rgb, 255:red, 208; green, 2; blue, 27 }  ,draw opacity=1 ]   (570.14,189.57) -- (620.14,189.57) ;

\draw (37,102) node  [align=left] {{\footnotesize 0.0}};
\draw (37,35) node  [align=left] {{\footnotesize 1.0}};
\draw (37,70) node  [align=left] {{\footnotesize 0.5}};
\draw (37,252) node  [align=left] {{\footnotesize 0.0}};
\draw (37,185) node  [align=left] {{\footnotesize 1.0}};
\draw (37,220) node  [align=left] {{\footnotesize 0.5}};
\draw (377,252) node  [align=left] {{\footnotesize 0.0}};
\draw (377,185) node  [align=left] {{\footnotesize 1.0}};
\draw (377,220) node  [align=left] {{\footnotesize 0.5}};
\draw (377,102) node  [align=left] {{\footnotesize 0.0}};
\draw (377,35) node  [align=left] {{\footnotesize 1.0}};
\draw (377,70) node  [align=left] {{\footnotesize 0.5}};
\draw (81,122) node  [align=left] {{\footnotesize 1.0}};
\draw (141,122) node  [align=left] {{\footnotesize 3.0}};
\draw (170,122) node  [align=left] {{\footnotesize 4.0}};
\draw (200,122) node  [align=left] {{\footnotesize 5.0}};
\draw (231,122) node  [align=left] {{\footnotesize 6.0}};
\draw (260,122) node  [align=left] {{\footnotesize 7.0}};
\draw (81,270) node  [align=left] {{\footnotesize 10.0}};
\draw (112,270) node  [align=left] {{\footnotesize 12.0}};
\draw (141,270) node  [align=left] {{\footnotesize 14.0}};
\draw (170,270) node  [align=left] {{\footnotesize 16.0}};
\draw (200,270) node  [align=left] {{\footnotesize 18.0}};
\draw (231,270) node  [align=left] {{\footnotesize 20.0}};
\draw (260,270) node  [align=left] {{\footnotesize 22.0}};
\draw (421,122) node  [align=left] {{\footnotesize 2.0}};
\draw (452,122) node  [align=left] {{\footnotesize 4.0}};
\draw (481,122) node  [align=left] {{\footnotesize 6.0}};
\draw (510,122) node  [align=left] {{\footnotesize 8.0}};
\draw (540,122) node  [align=left] {{\footnotesize 10.0}};
\draw (571,122) node  [align=left] {{\footnotesize 12.0}};
\draw (600,122) node  [align=left] {{\footnotesize 14.0}};
\draw (421,270) node  [align=left] {{\footnotesize 20.0}};
\draw (452,270) node  [align=left] {{\footnotesize 21.0}};
\draw (481,270) node  [align=left] {{\footnotesize 22.0}};
\draw (510,270) node  [align=left] {{\footnotesize 23.0}};
\draw (540,270) node  [align=left] {{\footnotesize 24.0}};
\draw (571,270) node  [align=left] {{\footnotesize 25.0}};
\draw (600,270) node  [align=left] {{\footnotesize 26.0}};
\draw (173,140) node [scale=0.9] [align=left] {$\displaystyle pres< 3.0\leftarrow $RampDown(3.0,5.0)};
\draw (111,122) node  [align=left] {{\footnotesize 2.0}};
\draw (172,290) node [scale=0.9] [align=left] {$\displaystyle pres >20.0\leftarrow $RampUp(15.0,20.0)};
\draw (514,140) node [scale=0.9] [align=left] {$\displaystyle lum< 5.0\leftarrow $RampDown(5.0,10.0)};
\draw (514,290) node [scale=0.9] [align=left] {$\displaystyle lum >25.0\leftarrow $RampUp(23.0,25.0)};

\draw (114,170) node  [align=left] {$\displaystyle \sqcap _{x_{2}\rightarrow x_{2}}$,$\displaystyle \sqcap _{x_{1}\rightarrow x_{2}}$};
\draw (413,21) node  [align=left] {$\displaystyle \sqcap _{x_{1}}$$ $};
\draw (413,170) node  [align=left] {$\displaystyle \sqcap _{x_{2}}$$ $};
\draw (114,21) node  [align=left] {$\displaystyle \sqcap _{x_{1}\rightarrow x_{1}}$,$\displaystyle \sqcap _{x_{2}\rightarrow x_{1}}$};

\end{tikzpicture}
\caption{State transition and state emission constraints defined as distributions of possibility for the model described in Fig.\ref{example101}.}
\label{example102}
\end{figure}
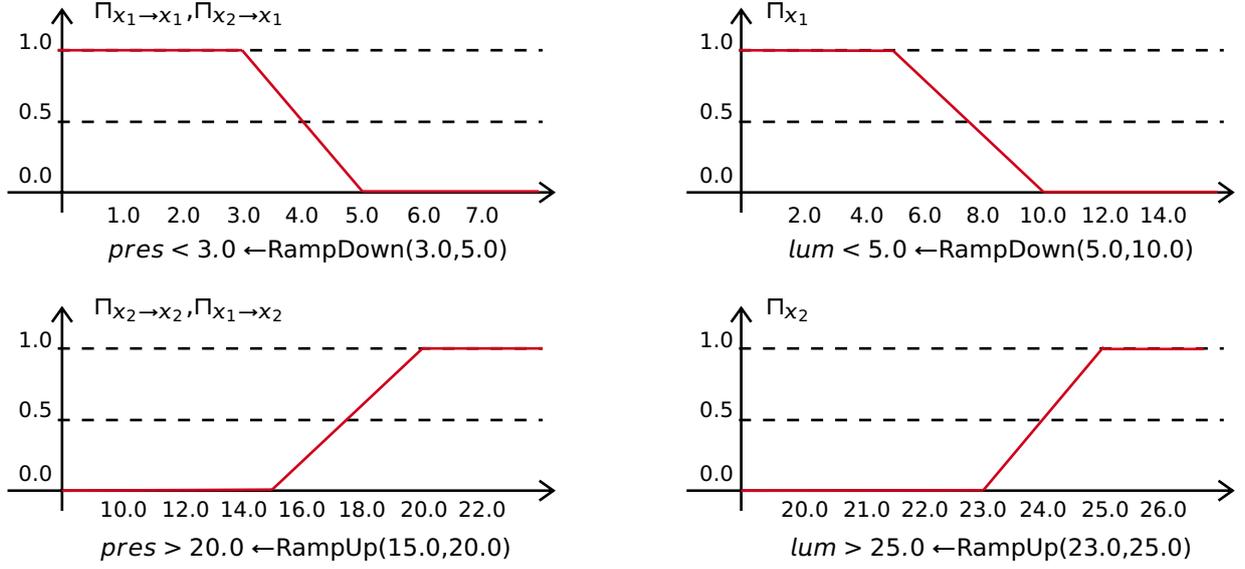

\noindent Given this model, let us now detail the basic mechanics of the Ev-IOHMM state-prediction. Let us consider the two states model depicted in Fig.\ref{example101} extending the model depicted in Fig.\ref{_dfsa_bis} with constraints taking into account uncertainties described as distributions of possibility as depicted in Fig.\ref{example102}.


\medskip

\noindent Let us assume that the input $pres$ at time $t-1$ was 3.5. Recall that constraints are encoded in the form of probability or possibility functions. So, one needs to compute the possibility value of the input value for each state-transition in the matrix $\mathbm{A'}$. It gives:
\medskip

\medskip

\begin{minipage}{0.972\linewidth}	
	\centering	
		\begin{tabular}{rcc}
			\hline
			& $\mathbf{x_{1_{(t)}}}$ & $\mathbf{x_{2_{(t)}}}$ \\
			\hline
			$x_{1_{(t-1)}}$ &  0.75  & 0.0 \\
			$x_{2_{(t-1)}}$ &  0.75  & 0.0 \\
			\hline									
		\end{tabular}
		\captionof{table}{\label{singletons}Possibility values at time \texorpdfstring{$t$}{} computed from matrix \texorpdfstring{$\mathbm{A'}$}{} when \texorpdfstring{$pres_{(t-1)}=3.5$}{}.}	
	\end{minipage}
\medskip

\noindent Now, beliefs allocated to the subsets $\in 2^\Omega$, i.e. elements of the matrix \textbf{A}, can be deduced from beliefs on the singletons obtained from the observations and the distributions of possibility described in matrix \textbf{A'} as follows.
\begin{itemize}
\item [--] When beliefs on singletons are obtained from probability density functions (likelihoods $\mathcal{L}(x_i|\vec{o})$), one can obtain commonality $q^\Omega$ by \cite{delmotte2004target}:
\begin{equation}\label{l}
q^\Omega[\vec{o}](A)=\prod_{x_i\in A}\mathcal{L}(x_i|\vec{o}), \forall A\in 2^\Omega
\end{equation}
\item [--] When beliefs on singletons are obtained from possibility distributions $\sqcap_{x_i}$, one can obtain plausibility $pl^\Omega$ by \cite{dubois2001new}:
\begin{flalign}
	pl^\Omega[\vec{o}](\{x_i\})&=\sqcap_{x_i}(\vec{o}) \label{poseq1}\\ 
	pl^\Omega[\vec{o}](A)&=\max_{x_i\in A}pl^\Omega[\vec{o}](\{x_i\}) \text{, } A \in 2^\Omega \label{poseq2} 
\end{flalign}
\end{itemize}

\noindent By applying Eq.\ref{poseq2} and then Eq.\ref{pm} for transforming $pl^\Omega$ to $m^\Omega$, one obtains:
\medskip

\medskip

\begin{minipage}{0.5\linewidth}	
	\centering	
		\begin{tabular}{rcccc}
			\hline
			& $\mathbf{\emptyset_{(t)}}$ & $\mathbf{\{x_{1_{(t)}}\}}$ & $\mathbf{\{x_{2_{(t)}}\}}$ & $\mathbf{\Omega_{(t)}}$\\
			\hline
			$[\emptyset_{(t-1)}]$ & ? & ?  & ? & ?\\		
			$[x_{1_{(t-1)}}]$ &  0.0 & 0.75  & 0.0  & 0.75\\
			$[x_{2_{(t-1)}}]$ &  0.0 & 0.75  & 0.0  & 0.75\\
			$[\Omega_{(t-1)}]$ &  ? & ?  & ?  & ?\\	
			\hline									
		\end{tabular}
		\captionof{table}{\label{pl}$pl^{\Omega_{(t)}}$ computed from Table.\ref{singletons} by applying Eq.\ref{poseq2}.}		
	\end{minipage}
\medskip
\hfill
\begin{minipage}{0.5\linewidth}	
	\centering	
		\begin{tabular}{rcccc}
			\hline
			& $\mathbf{\emptyset_{(t)}}$ & $\mathbf{\{x_{1_{(t)}}\}}$ & $\mathbf{\{x_{2_{(t)}}\}}$ & $\mathbf{\Omega_{(t)}}$\\
			\hline
			$[\emptyset_{(t-1)}]$ & ? & ?  & ? & ?\\		
			$[x_{1_{(t-1)}}]$ &  0.25 & 0.75  & 0.0  & 0.0\\
			$[x_{2_{(t-1)}}]$ &  0.25 & 0.75  & 0.0  & 0.0\\
			$[\Omega_{(t-1)}]$ &  ? & ?  & ?  & ?\\	
			\hline									
		\end{tabular}
		\captionof{table}{\label{bbas}$m^{\Omega_{(t)}}$ computed from Table.\ref{pl} by applying Eq.\ref{pm}.}		
	\end{minipage}
\medskip

\noindent For the time being, only the BBAs conditional to the singletons $[x_{1_{(t-1)}}]$ and $[x_{2_{(t-1)}}]$ are available. So, one needs to compute beliefs conditional to the subsets $[\emptyset_{(t-1)}]$ and $[\Omega_{(t-1)}]$. Per \cite{smets2008belief}, this can be achieved by applying a DRC on BBAs conditional to the singletons as follows:
\begin{equation}\label{extension}
m^{\Omega_{(t)}}_a[A] = \drc_{x_i\in A} m^{\Omega_{(t)}}_a[x_i] \text{, } \forall A\in 2^{\Omega_{(t-1)}}
\end{equation}
\noindent For instance, $m^{\Omega_{(t)}}_a[\Omega]= m^{\Omega_{(t)}}_a[x_{1_{(t-1)}}]\drc m^{\Omega_{(t)}}_a[x_{2_{(t-1)}}]$. Results are given in Table.\ref{tme}.
\medskip

\begin{minipage}{0.972\linewidth}	
	\centering	
	\begin{tabular}{rcccc}
		\hline
		& $\mathbf{\emptyset_{(t)}}$ & $\mathbf{\{x_{1_{(t)}}\}}$ & $\mathbf{\{x_{2_{(t)}}\}}$ & $\mathbf{\Omega_{(t)}}$\\
		\hline
		$[\emptyset_{(t-1)}]$ & 1 & 0  & 0 & 0\\		
		$[\{x_{1_{(t-1)}}\}]$ & 0.25 & 0.75  & 0.0  & 0.0\\
		$[\{x_{2_{(t-1)}}\}]$ & 0.25 & 0.75  & 0.0  & 0.0\\										
		$[\Omega_{(t-1)}]$ &  0.0625 & 0.9375  & 0.0  & 0.0\\
		\hline
	\end{tabular}	
	\captionof{table}{\label{tme}State-transition matrix computed from Table.\ref{bbas} after application of Eq.\ref{extension}.}	
\end{minipage}
\medskip

\noindent We are now ready to compute states prediction at time $t$ given states at time $t-1$. The prediction is obtained using the following generalized conjunctive form \cite{smets2008belief}:
\begin{equation}\label{f}
\boxed{
\hat{q}^{\Omega_{(t)}}_\alpha(A) =\sum_{X \in 2^{\Omega_{(t-1)}}}m^{\Omega_{(t-1)}}_\alpha(X)\cdot  q^{\Omega_{(t)}}_a[X,\vec{u}_{(t-1)}](A) \text{, } \forall A\in 2^{\Omega_{(t)}}, X\in 2^{\Omega_{(t-1)}}
}
\end{equation}
\noindent where $q^{\Omega_{(t)}}_a[X,\vec{u}_{(t-1)}](A)$ corresponds to the matrix given in Table.\ref{tme}. Without an a priori on the previous states, i.e. $m^{\Omega_{(t-1)}}_\alpha(\Omega_{(t-1)})=1$, the predicted BBA is given from Eq.\ref{f} further transformed to $m$:
\medskip 

\begin{minipage}{0.972\linewidth}	
	\centering	
		\begin{tabular}{rcccc}
			\hline
			& $\mathbf{\emptyset_{(t)}}$ & $\mathbf{\{x_{1_{(t)}}\}}$ & $\mathbf{\{x_{2_{(t)}}\}}$ & $\mathbf{\Omega_{(t)}}$\\
			\hline
			$\hat{m}^{\Omega_{(t)}}_\alpha$ & 0.25 & 0.75  & 0  & 0\\
			\hline									
		\end{tabular}
		\captionof{table}{\label{estimatedBBA}State estimation at time $t$ given $pres_{(t-1)}=3.5$.}	
	\end{minipage}
\medskip

\subsubsection{State emission}
\label{se}
\noindent Let us also assume that the output $lum$ at time $t$ is 2.34. So, one needs to compute the possibility value of the output value for each state from the vector $\mathbm{\vec{B}'}$ from which the BBA can further be computed. It gives: 
\medskip

\medskip

\begin{minipage}{0.972\linewidth}	
	\centering	
		\begin{tabular}{cc}
			\hline
			$\mathbf{x_{1_{(t)}}}$ & $\mathbf{x_{2_{(t)}}}$ \\
			\hline
			1.0  & 0.0 \\
			\hline									
		\end{tabular}
		\captionof{table}{\label{ke}Possibility values at time $t$ computed from vector $\mathbm{\vec{B}'}$ when $lum_{(t)}=2.34$.}	
	\end{minipage}
\medskip

\medskip

\noindent By applying Eq.\ref{poseq2} and then Eq.\ref{pm} for transforming $pl^\Omega$ to $m^\Omega$, one obtains:
\medskip

\medskip

\begin{minipage}{0.5\linewidth}	
	\centering	
		\begin{tabular}{rcccc}
			\hline
			$\mathbf{\emptyset_{(t)}}$ & $\mathbf{\{x_{1_{(t)}}\}}$ & $\mathbf{\{x_{2_{(t)}}\}}$ & $\mathbf{\Omega_{(t)}}$\\
			\hline
			0.0 & 1.0  & 0.0  & 1.0\\
			\hline									
		\end{tabular}
		\captionof{table}{\label{pl2}$pl^{\Omega_{(t)}}$ computed from Table.\ref{ke} by applying Eq.\ref{poseq2}.}		
	\end{minipage}
\medskip
\hfill
\begin{minipage}{0.5\linewidth}	
	\centering	
		\begin{tabular}{rcccc}
			\hline
			& $\mathbf{\emptyset_{(t)}}$ & $\mathbf{\{x_{1_{(t)}}\}}$ & $\mathbf{\{x_{2_{(t)}}\}}$ & $\mathbf{\Omega_{(t)}}$\\
			\hline
			$K_E$ & 0.0 & 1.0  & 0.0  & 0.0\\
			\hline									
		\end{tabular}
		\captionof{table}{\label{__bbas}$m^{\Omega_{(t)}}$ computed from Table.\ref{pl2} by applying Eq.\ref{pm}.}		
	\end{minipage}
\medskip

\noindent In the sequel, the Ev-IOHMM forward algorithm is detailed. This algorithm computes the likelihood of the observation sequences $(\vec{u}_t,\vec{y}_t)_{t=1}^T$ in the form of a BBA from which the degree of effectiveness is computed. 

\begin{note}
The forward algorithm described in the next section makes use of the CRC (\ref{__crc}) for propagating beliefs. Other combination rules such as the Cautious Conjunctive Rule of Combination (CCRC) and the Bold Disjunctive Rule of Combination (BDRC) have been introduced in the TBM framework \cite{denoeux2008conjunctive}. \textbf{However, it is shown that the CRC is the only rule satisfying the Shafer-Shenoy axioms for belief functions propagation \cite{pichon2009fonctions}\cite{shenoy2008axioms}.}
\end{note}

\subsubsection{The Ev-IOHMM Forward algorithm}
\label{evfa}
\noindent The Ev-IOHMM forward algorithm is close to the Ev-HMM forward algorithm described in \cite{ramasso2017inference} and \cite{serir2011time}. The main difference consists in conditioning the state-transition not only on the previous state $x_{(t-1)}$ but also on the input observation $\vec{u}_{(t-1)}$. Thus, the forward algorithm is given by : 

\medskip
\noindent \textbf{Initialization}\\
\noindent no a priori is given to the initial state of the system, i.e. $m^\Omega_{pi}(\Omega)=1$. Thus, $\forall X_{(t=1)}\in 2^\Omega$
\begin{equation}
m^{\Omega_{(t=1)}}_\alpha(X_{(t=1)})=\overbrace{m^{\Omega_{(t=1)}}_b[\vec{y}_{(t=1)}](X_{(t=1)})}^{K_E\textbf{ (State emission)}}
\end{equation}

\noindent \textbf{Induction}\\
\noindent $\forall X_{(t)}\in 2^\Omega$, $2\leq t \leq T$
\medskip

\begin{equation}
q^{\Omega_{(t)}}_\alpha(X_{(t)})=\overbrace{q^{\Omega_{(t)}}_b[\vec{y}_{(t)}](X_{(t)})}^{K_E\textbf{ (State emission)}}\cdot\overbrace{\sum_{X_i\in 2^{\Omega_{(t-1)}}}m^{\Omega_{(t-1)}}_\alpha(X_i)\cdot q^{\Omega_{(t)}}_a[X_i,\vec{u}_{(t-1)}](X_{(t)})
}^{K_S\textbf{ (State prediction)}}
\end{equation}
\medskip

\noindent It is worth noting that, at each time $t$, the resulting $q^{\Omega_{(t)}}_\alpha(X_{(t)})$ has to be transformed to $m^{\Omega_{(t)}}_\alpha(X_{(t)})$ by using Eq.\ref{qm}. Thus, one obtain the BBA $m^{\Omega_{(t)}}_\alpha$ resulting from the combination of the belief at time $t$ with the previous beliefs combined together. However, successive combinations lead the conflict to increase over time, i.e.
\begin{equation}
	\lim_{T\to\infty}\mathbf{m^{\Omega_{(T)}}_\alpha(\emptyset)}=1.0
\end{equation}

\noindent So as to cope with this problem, the BBA $m^{\Omega_{(t)}}_\alpha$ is normalized at each time $t$ (from $t=1$) by redistributing the conflict over propositions $\in 2^\Omega$. Several strategies have been defined for redistributing the conflict \cite{martin2008general} so as to keep $m^{\Omega_{(t)}}_\alpha(\emptyset)=0$ at each time $t$. For instance, assuming sources are equally reliable and ${m_{1\crc 2}^\Omega}(\emptyset) < 1$, the Dempster's normalization rule redistributes conflict on the focal elements, i.e. $\forall A\in 2^\Omega, A\neq\emptyset$ and ${m_{1\crc 2}^\Omega}(A)>0$,
\begin{equation}
	m_D^\Omega(A)=\frac{1}{1-{m_{1\crc 2}^\Omega}(\emptyset)}\cdot
	{m_{1\crc 2}^\Omega}(A) \text{,   } {m_{1\crc 2}^\Omega}(\emptyset)=0
\end{equation}	
\noindent Other normalization rules exist \cite{martin2008general}. For instance, Dubois-Prade normalization rule assumes that at least one source is reliable in case of conflict. $\forall A\in 2^\Omega, A\neq\emptyset$, the rule is defined by:
\begin{equation}\label{dp}
	m_{DP}^\Omega(A)= {m_{1\crc 2}^\Omega}(A) + \sum_{B\cup C=A, B\cap C=\emptyset} m_1^\Omega(B)\cdot m_2^\Omega(C)
\end{equation}

\noindent \textbf{Termination}:\\
\noindent In this paper, we do consider leveraging the conflict as a means to provide the degree of effectiveness as done in \cite{ramasso2017inference}. Following the proof given in \cite{ramasso2017inference} in the context of Ev-HMM where :

\begin{equation}\label{final}
pl^{\Omega_{(T)}}_\alpha(\Omega) = 1 - m^{\Omega_{(T)}}_\alpha(\emptyset)
\end{equation}

\noindent the degree of effectiveness given the Ev-IOHMM $\Theta$ and the observation sequence $(\vec{u}_t,\vec{y}_t)_{t=1}^T$ is given by :

\begin{equation}\label{_final}
	\boxed{
pl(\Theta,(\vec{u}_t,\vec{y}_t)_{t=1}^T)=\prod_{t=1}^T \left(1 - m^{\Omega_{(t)}}_\alpha(\emptyset)\right)}
\end{equation}

\noindent As described in \cite{serir2011time}, one needs to record the value of the conflict at each time $t$ before normalizing the BBA.

\begin{note}
\noindent The degree of effectiveness may also be a good indicator of the quality of the model \cite{serir2011time}. Indeed, given a perfect sequence of observations, the degree of effectiveness is supposed to be equal to 1.0 (see Definition.\ref{comfort}).
\end{note}
\section{An application to autonomous vehicles}
\label{app}
\noindent Autonomous vehicles are gaining momentum in the CPS community. Although promising a breakthrough in terms of traffic optimization and regulation, these vehicles will be unable to fulfil their potential without ensuring safety of passengers and surroundings. Besides the physical environment these vehicles operate in and in which unanticipated events may hamper their operation at any time, these vehicles are also prone to cyber attacks, potential communication infrastructure and electronic devices issues \cite{dutta2018security}. For designers, it is then about handling the behavior of these vehicles from an holistic point of view rather than considering each part of the system separately \cite{lima2016towards},\cite{katsikas2017cyber}. Thus, by modeling the expected behavior of the system taken as a whole rather than considering its internals, the proposed approach is coherent with the holistic point of view.\\
\medskip

\noindent This section aims at providing a possible application of the method developed throughout the paper to the domain of autonomous vehicles. The proposed application consists in providing autonomous vehicles designers with a bench-marking tool used for assessing the effectiveness of the controllers of these vehicles. Without claiming to be exhaustive, the solution can be used in addition to existing approaches for autonomous vehicles security and safety. Two scenarios are then considered. The first one is about considering the speed limitations in force in France. Such limitation rules are complex and depend on several factors depicted in Table.\ref{slf}.  

\begin{table}[ht!]
\small
\centering
\begin{tabular}{@{}lllll@{}}
\toprule
\textbf{Localization}                        & \textbf{Weather condition}        & \textbf{Road}    & \textbf{Improved road} & \textbf{Highway}                                                           \\ \midrule
\multirow{3}{*}{Outside urban area} & No precipitation         & 90 km/h & 110 km/h      & 130 km/h                                                          \\
                                    & Rainy                    & 80 km/h & 100 km/h      & 110 km/h                                                          \\
                                    & Visibility \textless 50m & 50 km/h & 50 km/h       & 50 km/h                                                           \\ \midrule
\multirow{3}{*}{Urban area}         & 30 km/h area             & 30 km/h & 30 km/h       &   N/A                                                                \\
                                    & General case             & 50 km/h & 50 km/h       & \begin{tabular}[c]{@{}l@{}}90 km/h\\ (urban highway)\end{tabular} \\
                                    & Improved section         & 70 km/h & 70 km/h       &   N/A                                                                \\ \cmidrule(l){1-5} 
\end{tabular}
\caption{\label{slf} Speed limitations in force and their dependency on several factors.}
\end{table}
\noindent The idea here is to prevent the vehicle going beyond the maximum speed allowed taking into account factors depicted in Table.\ref{slf}. The second scenario consists in considering passengers' well being. The idea here is to prevent rapid accelerations and decelerations of the vehicle especially when meeting humps, roundabout, "giveway", etc.

\subsection{Methodology}
\noindent OpenStreetMap (OSM) is a collaborative map of the world. It is a powerful source of information about all types of infrastructure features such as roads, trails, side-walks, etc. Specifically, given GPS longitude and latitude, one can retrieve from the repository all information about the infrastructure around. Our approach is to use this repository so as to gather infrastructure information from GPS data at each time $t$ as the vehicle moves. Besides OSM, we do use a weather web service (api.weatherbit.io) allowing us to retrieve weather information at the date of interest.\\

\noindent Based on these information, we do first describe the Ev-HMM model corresponding to the first scenario. Table.\ref{speed_evhmm} provides a summary of the features provided by OSM regarding the speed limitations and the associated speed limit values further used in the model. Besides speed limitations, the weather web service provides us with precipitation values (in mm) and visibility (in km). Speed limitation values and weather information are used as inputs of the Ev-IOHMM model (multivariate). Instant speed of the vehicle is used as output of the Ev-IOHMM model. All the constraints considered, the model contains 11 states. For the sake of visibility, a partial representation of the model is depicted in Fig.\ref{partial}.

\begin{table}[ht!]
\small
\centering
\begin{tabular}{@{}lllll@{}}
\toprule
\multicolumn{3}{l}{\textbf{OSM Road infrastructure max speed features}} & \multicolumn{2}{l}{\textbf{Weather data}} \\ \midrule
From values & \cons{maxspeed:<value>} & value & \cons{hour.precip} & mm \\
\multirow{12}{*}{From localization} & \cons{zone:maxspeed:FR:30} & 30 km/h & \cons{hour.vis} & km \\
 & \cons{maxspeed:type:FR:urban} & 50 km/h &  &  \\
 & \cons{maxspeed:type:FR:rural} & 90 km/h &  &  \\
 & \cons{maxspeed:type:FR:trunk} & 110 km/h &  &  \\
 & \cons{maxspeed:type:FR:motorway} & 130 km/h &  &  \\
 & \cons{highway:living\_street} & 50 km/h &  &  \\
 & \cons{highway:residential} & 50 km/h &  &  \\
 & \cons{highway:primary} & 90 km/h &  &  \\
 & \cons{highway:secondary} & 70 km/h &  &  \\
 & \cons{highway:tertiary} & 50 km/h &  &  \\
 & \cons{highway:trunk} & 110 km/h &  &  \\
 & \cons{highway:motorway} & 130 km/h &  &  \\ \cmidrule(l){1-5} 
\end{tabular}%
\caption{\label{speed_evhmm} Speed limitation values are retrieved at each time $t$ from OSM specific infrastructure features given the vehicle position (longitude/latitude). The weather web services provides us with precipitation and visibility information.}
\end{table}

\tikzset{every picture/.style={line width=0.75pt}} 
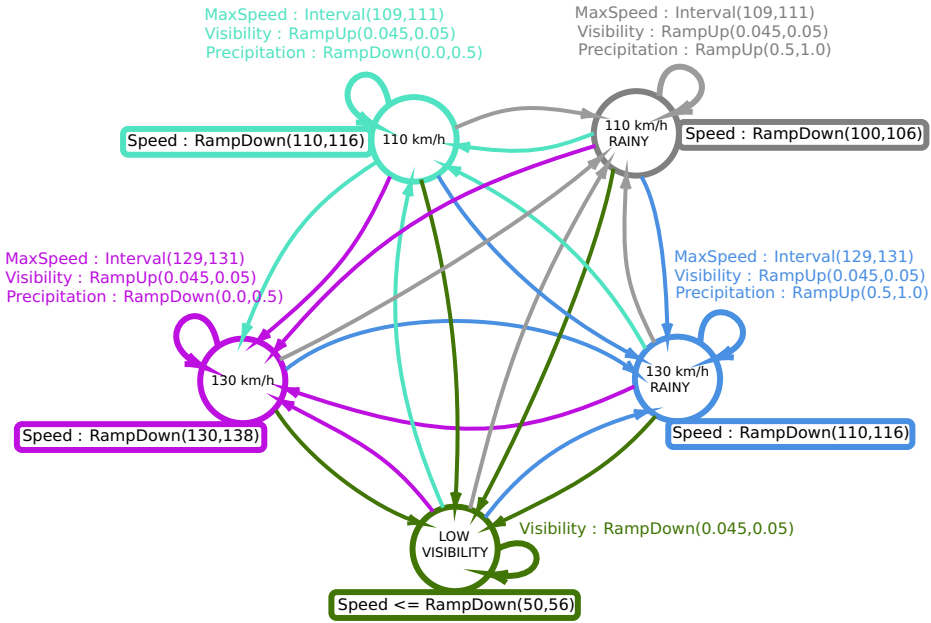
\begin{figure}[ht!]
\centering
\begin{tikzpicture}[scale=0.5, every node/.style={scale=0.5}, x=0.75pt,y=0.75pt,yscale=-1,xscale=1]
\draw  [color={rgb, 255:red, 189; green, 16; blue, 224 }  ,draw opacity=1 ][line width=2.25]  (203,541.5) .. controls (203,518.03) and (222.03,499) .. (245.5,499) .. controls (268.97,499) and (288,518.03) .. (288,541.5) .. controls (288,564.97) and (268.97,584) .. (245.5,584) .. controls (222.03,584) and (203,564.97) .. (203,541.5) -- cycle ;
\draw [color={rgb, 255:red, 189; green, 16; blue, 224 }  ,draw opacity=1 ][line width=2.25]    (220.5,506) .. controls (208.74,442.3) and (146.08,496.73) .. (201.93,525.29) ;
\draw [shift={(205.5,527)}, rotate = 204.3] [color={rgb, 255:red, 189; green, 16; blue, 224 }  ,draw opacity=1 ][line width=2.25]    (17.49,-5.26) .. controls (11.12,-2.23) and (5.29,-0.48) .. (0,0) .. controls (5.29,0.48) and (11.12,2.23) .. (17.49,5.26)   ;

\draw  [color={rgb, 255:red, 74; green, 144; blue, 226 }  ,draw opacity=1 ][line width=2.25]  (637,539.5) .. controls (637,516.03) and (656.03,497) .. (679.5,497) .. controls (702.97,497) and (722,516.03) .. (722,539.5) .. controls (722,562.97) and (702.97,582) .. (679.5,582) .. controls (656.03,582) and (637,562.97) .. (637,539.5) -- cycle ;
\draw [color={rgb, 255:red, 74; green, 144; blue, 226 }  ,draw opacity=1 ][line width=2.25]    (702.5,502) .. controls (703.49,438.96) and (784.03,492.35) .. (720.5,517.85) ;
\draw [shift={(717.5,519)}, rotate = 340.08000000000004] [color={rgb, 255:red, 74; green, 144; blue, 226 }  ,draw opacity=1 ][line width=2.25]    (17.49,-5.26) .. controls (11.12,-2.23) and (5.29,-0.48) .. (0,0) .. controls (5.29,0.48) and (11.12,2.23) .. (17.49,5.26)   ;

\draw [color={rgb, 255:red, 74; green, 144; blue, 226 }  ,draw opacity=1 ][line width=1.5]    (288.5,530) .. controls (375.07,459.35) and (559.64,469.89) .. (637.34,533.04) ;
\draw [shift={(638.5,534)}, rotate = 219.73] [color={rgb, 255:red, 74; green, 144; blue, 226 }  ,draw opacity=1 ][line width=1.5]    (14.21,-4.28) .. controls (9.04,-1.82) and (4.3,-0.39) .. (0,0) .. controls (4.3,0.39) and (9.04,1.82) .. (14.21,4.28)   ;

\draw [color={rgb, 255:red, 189; green, 16; blue, 224 }  ,draw opacity=1 ][line width=1.5]    (636.5,547) .. controls (518.09,602.72) and (446.22,603.99) .. (289.86,549.82) ;
\draw [shift={(287.5,549)}, rotate = 379.19] [color={rgb, 255:red, 189; green, 16; blue, 224 }  ,draw opacity=1 ][line width=1.5]    (14.21,-4.28) .. controls (9.04,-1.82) and (4.3,-0.39) .. (0,0) .. controls (4.3,0.39) and (9.04,1.82) .. (14.21,4.28)   ;

\draw  [color={rgb, 255:red, 65; green, 117; blue, 5 }  ,draw opacity=1 ][line width=2.25]  (415,710.5) .. controls (415,687.03) and (434.03,668) .. (457.5,668) .. controls (480.97,668) and (500,687.03) .. (500,710.5) .. controls (500,733.97) and (480.97,753) .. (457.5,753) .. controls (434.03,753) and (415,733.97) .. (415,710.5) -- cycle ;
\draw [color={rgb, 255:red, 65; green, 117; blue, 5 }  ,draw opacity=1 ][line width=2.25]    (500,710.5) .. controls (544.82,686.37) and (560.04,747.13) .. (495.5,736.53) ;
\draw [shift={(492.5,736)}, rotate = 370.82] [color={rgb, 255:red, 65; green, 117; blue, 5 }  ,draw opacity=1 ][line width=2.25]    (17.49,-5.26) .. controls (11.12,-2.23) and (5.29,-0.48) .. (0,0) .. controls (5.29,0.48) and (11.12,2.23) .. (17.49,5.26)   ;

\draw [color={rgb, 255:red, 65; green, 117; blue, 5 }  ,draw opacity=1 ][line width=1.5]    (657.5,577) .. controls (610.97,631.45) and (568.36,651.6) .. (495.71,686.92) ;
\draw [shift={(493.5,688)}, rotate = 334.06] [color={rgb, 255:red, 65; green, 117; blue, 5 }  ,draw opacity=1 ][line width=1.5]    (14.21,-4.28) .. controls (9.04,-1.82) and (4.3,-0.39) .. (0,0) .. controls (4.3,0.39) and (9.04,1.82) .. (14.21,4.28)   ;

\draw [color={rgb, 255:red, 74; green, 144; blue, 226 }  ,draw opacity=1 ][line width=1.5]    (487.5,679) .. controls (528.09,624.55) and (577.5,591.66) .. (648.35,572.57) ;
\draw [shift={(650.5,572)}, rotate = 525.22] [color={rgb, 255:red, 74; green, 144; blue, 226 }  ,draw opacity=1 ][line width=1.5]    (14.21,-4.28) .. controls (9.04,-1.82) and (4.3,-0.39) .. (0,0) .. controls (4.3,0.39) and (9.04,1.82) .. (14.21,4.28)   ;

\draw [color={rgb, 255:red, 65; green, 117; blue, 5 }  ,draw opacity=1 ][line width=1.5]    (278.5,571) .. controls (309.86,620.98) and (376.75,670) .. (421.77,684.18) ;
\draw [shift={(424.5,685)}, rotate = 196.11] [color={rgb, 255:red, 65; green, 117; blue, 5 }  ,draw opacity=1 ][line width=1.5]    (14.21,-4.28) .. controls (9.04,-1.82) and (4.3,-0.39) .. (0,0) .. controls (4.3,0.39) and (9.04,1.82) .. (14.21,4.28)   ;

\draw [color={rgb, 255:red, 189; green, 16; blue, 224 }  ,draw opacity=1 ][line width=1.5]    (435.5,673) .. controls (390.95,614.59) and (360.12,602.24) .. (284.8,562.22) ;
\draw [shift={(282.5,561)}, rotate = 388.03] [color={rgb, 255:red, 189; green, 16; blue, 224 }  ,draw opacity=1 ][line width=1.5]    (14.21,-4.28) .. controls (9.04,-1.82) and (4.3,-0.39) .. (0,0) .. controls (4.3,0.39) and (9.04,1.82) .. (14.21,4.28)   ;

\draw  [color={rgb, 255:red, 80; green, 227; blue, 194 }  ,draw opacity=1 ][line width=2.25]  (374,298.5) .. controls (374,275.03) and (393.03,256) .. (416.5,256) .. controls (439.97,256) and (459,275.03) .. (459,298.5) .. controls (459,321.97) and (439.97,341) .. (416.5,341) .. controls (393.03,341) and (374,321.97) .. (374,298.5) -- cycle ;
\draw [color={rgb, 255:red, 80; green, 227; blue, 194 }  ,draw opacity=1 ][line width=2.25]    (391.5,263) .. controls (379.74,199.3) and (317.08,253.73) .. (372.93,282.29) ;
\draw [shift={(376.5,284)}, rotate = 204.3] [color={rgb, 255:red, 80; green, 227; blue, 194 }  ,draw opacity=1 ][line width=2.25]    (17.49,-5.26) .. controls (11.12,-2.23) and (5.29,-0.48) .. (0,0) .. controls (5.29,0.48) and (11.12,2.23) .. (17.49,5.26)   ;

\draw [color={rgb, 255:red, 189; green, 16; blue, 224 }  ,draw opacity=1 ][line width=1.5]    (392.5,336) .. controls (359.83,409.26) and (336.96,437.44) .. (261.79,500.09) ;
\draw [shift={(259.5,502)}, rotate = 320.27] [color={rgb, 255:red, 189; green, 16; blue, 224 }  ,draw opacity=1 ][line width=1.5]    (14.21,-4.28) .. controls (9.04,-1.82) and (4.3,-0.39) .. (0,0) .. controls (4.3,0.39) and (9.04,1.82) .. (14.21,4.28)   ;

\draw [color={rgb, 255:red, 80; green, 227; blue, 194 }  ,draw opacity=1 ][line width=1.5]    (379.5,321) .. controls (316.81,361.8) and (283.83,395.66) .. (246.07,496.48) ;
\draw [shift={(245.5,498)}, rotate = 290.43] [color={rgb, 255:red, 80; green, 227; blue, 194 }  ,draw opacity=1 ][line width=1.5]    (14.21,-4.28) .. controls (9.04,-1.82) and (4.3,-0.39) .. (0,0) .. controls (4.3,0.39) and (9.04,1.82) .. (14.21,4.28)   ;

\draw [color={rgb, 255:red, 65; green, 117; blue, 5 }  ,draw opacity=1 ][line width=1.5]    (423.5,339) .. controls (457.33,458.4) and (464.43,552.06) .. (457.6,666.28) ;
\draw [shift={(457.5,668)}, rotate = 273.48] [color={rgb, 255:red, 65; green, 117; blue, 5 }  ,draw opacity=1 ][line width=1.5]    (14.21,-4.28) .. controls (9.04,-1.82) and (4.3,-0.39) .. (0,0) .. controls (4.3,0.39) and (9.04,1.82) .. (14.21,4.28)   ;

\draw [color={rgb, 255:red, 80; green, 227; blue, 194 }  ,draw opacity=1 ][line width=1.5]    (445.5,669) .. controls (395.75,553.58) and (384.61,462.91) .. (413.07,342.81) ;
\draw [shift={(413.5,341)}, rotate = 463.48] [color={rgb, 255:red, 80; green, 227; blue, 194 }  ,draw opacity=1 ][line width=1.5]    (14.21,-4.28) .. controls (9.04,-1.82) and (4.3,-0.39) .. (0,0) .. controls (4.3,0.39) and (9.04,1.82) .. (14.21,4.28)   ;

\draw  [color={rgb, 255:red, 128; green, 128; blue, 128 }  ,draw opacity=1 ][line width=2.25]  (596,292.5) .. controls (596,269.03) and (615.03,250) .. (638.5,250) .. controls (661.97,250) and (681,269.03) .. (681,292.5) .. controls (681,315.97) and (661.97,335) .. (638.5,335) .. controls (615.03,335) and (596,315.97) .. (596,292.5) -- cycle ;
\draw [color={rgb, 255:red, 155; green, 155; blue, 155 }  ,draw opacity=1 ][line width=2.25]    (661.5,255) .. controls (662.49,191.96) and (743.03,245.35) .. (679.5,270.85) ;
\draw [shift={(676.5,272)}, rotate = 340.08000000000004] [color={rgb, 255:red, 155; green, 155; blue, 155 }  ,draw opacity=1 ][line width=2.25]    (17.49,-5.26) .. controls (11.12,-2.23) and (5.29,-0.48) .. (0,0) .. controls (5.29,0.48) and (11.12,2.23) .. (17.49,5.26)   ;

\draw [color={rgb, 255:red, 155; green, 155; blue, 155 }  ,draw opacity=1 ][line width=1.5]    (656.5,503) .. controls (626.8,443.6) and (623.56,423.4) .. (628.35,336.65) ;
\draw [shift={(628.5,334)}, rotate = 453.22] [color={rgb, 255:red, 155; green, 155; blue, 155 }  ,draw opacity=1 ][line width=1.5]    (14.21,-4.28) .. controls (9.04,-1.82) and (4.3,-0.39) .. (0,0) .. controls (4.3,0.39) and (9.04,1.82) .. (14.21,4.28)   ;

\draw [color={rgb, 255:red, 74; green, 144; blue, 226 }  ,draw opacity=1 ][line width=1.5]    (643.5,336) .. controls (660.33,365.7) and (668.34,423.82) .. (669.47,495.81) ;
\draw [shift={(669.5,498)}, rotate = 269.22] [color={rgb, 255:red, 74; green, 144; blue, 226 }  ,draw opacity=1 ][line width=1.5]    (14.21,-4.28) .. controls (9.04,-1.82) and (4.3,-0.39) .. (0,0) .. controls (4.3,0.39) and (9.04,1.82) .. (14.21,4.28)   ;

\draw [color={rgb, 255:red, 155; green, 155; blue, 155 }  ,draw opacity=1 ][line width=1.5]    (472.5,670) .. controls (506.33,531.69) and (532.24,453.78) .. (608.35,325.93) ;
\draw [shift={(609.5,324)}, rotate = 480.83] [color={rgb, 255:red, 155; green, 155; blue, 155 }  ,draw opacity=1 ][line width=1.5]    (14.21,-4.28) .. controls (9.04,-1.82) and (4.3,-0.39) .. (0,0) .. controls (4.3,0.39) and (9.04,1.82) .. (14.21,4.28)   ;

\draw [color={rgb, 255:red, 65; green, 117; blue, 5 }  ,draw opacity=1 ][line width=1.5]    (615.5,328) .. controls (606.54,385.71) and (557.99,524.6) .. (480.67,670.8) ;
\draw [shift={(479.5,673)}, rotate = 297.95] [color={rgb, 255:red, 65; green, 117; blue, 5 }  ,draw opacity=1 ][line width=1.5]    (14.21,-4.28) .. controls (9.04,-1.82) and (4.3,-0.39) .. (0,0) .. controls (4.3,0.39) and (9.04,1.82) .. (14.21,4.28)   ;

\draw [color={rgb, 255:red, 155; green, 155; blue, 155 }  ,draw opacity=1 ][line width=1.5]    (457.5,287) .. controls (507.74,264.34) and (530.8,262.06) .. (594.56,274.43) ;
\draw [shift={(597.5,275)}, rotate = 191.14] [color={rgb, 255:red, 155; green, 155; blue, 155 }  ,draw opacity=1 ][line width=1.5]    (14.21,-4.28) .. controls (9.04,-1.82) and (4.3,-0.39) .. (0,0) .. controls (4.3,0.39) and (9.04,1.82) .. (14.21,4.28)   ;

\draw [color={rgb, 255:red, 80; green, 227; blue, 194 }  ,draw opacity=1 ][line width=1.5]    (596,292.5) .. controls (552.44,309.33) and (529.46,315.38) .. (460.12,304.82) ;
\draw [shift={(458,304.5)}, rotate = 368.81] [color={rgb, 255:red, 80; green, 227; blue, 194 }  ,draw opacity=1 ][line width=1.5]    (14.21,-4.28) .. controls (9.04,-1.82) and (4.3,-0.39) .. (0,0) .. controls (4.3,0.39) and (9.04,1.82) .. (14.21,4.28)   ;

\draw [color={rgb, 255:red, 74; green, 144; blue, 226 }  ,draw opacity=1 ][line width=1.5]    (440.5,335) .. controls (457.42,364.85) and (517.89,455.09) .. (639.66,518.05) ;
\draw [shift={(641.5,519)}, rotate = 207.12] [color={rgb, 255:red, 74; green, 144; blue, 226 }  ,draw opacity=1 ][line width=1.5]    (14.21,-4.28) .. controls (9.04,-1.82) and (4.3,-0.39) .. (0,0) .. controls (4.3,0.39) and (9.04,1.82) .. (14.21,4.28)   ;

\draw [color={rgb, 255:red, 80; green, 227; blue, 194 }  ,draw opacity=1 ][line width=1.5]    (647.5,509) .. controls (598.99,419.9) and (532.84,366.08) .. (451.47,326.69) ;
\draw [shift={(449,325.5)}, rotate = 385.58000000000004] [color={rgb, 255:red, 80; green, 227; blue, 194 }  ,draw opacity=1 ][line width=1.5]    (14.21,-4.28) .. controls (9.04,-1.82) and (4.3,-0.39) .. (0,0) .. controls (4.3,0.39) and (9.04,1.82) .. (14.21,4.28)   ;

\draw  [color={rgb, 255:red, 189; green, 16; blue, 224 }  ,draw opacity=1 ][line width=2.25]  (20,590.2) .. controls (20,587.33) and (22.33,585) .. (25.2,585) -- (261.8,585) .. controls (264.67,585) and (267,587.33) .. (267,590.2) -- (267,605.8) .. controls (267,608.67) and (264.67,611) .. (261.8,611) -- (25.2,611) .. controls (22.33,611) and (20,608.67) .. (20,605.8) -- cycle ;
\draw  [color={rgb, 255:red, 80; green, 227; blue, 194 }  ,draw opacity=1 ][line width=2.25]  (126,293.2) .. controls (126,290.33) and (128.33,288) .. (131.2,288) -- (368.3,288) .. controls (371.17,288) and (373.5,290.33) .. (373.5,293.2) -- (373.5,308.8) .. controls (373.5,311.67) and (371.17,314) .. (368.3,314) -- (131.2,314) .. controls (128.33,314) and (126,311.67) .. (126,308.8) -- cycle ;
\draw  [color={rgb, 255:red, 128; green, 128; blue, 128 }  ,draw opacity=1 ][line width=2.25]  (681,286.2) .. controls (681,283.33) and (683.33,281) .. (686.2,281) -- (922.8,281) .. controls (925.67,281) and (928,283.33) .. (928,286.2) -- (928,301.8) .. controls (928,304.67) and (925.67,307) .. (922.8,307) -- (686.2,307) .. controls (683.33,307) and (681,304.67) .. (681,301.8) -- cycle ;
\draw  [color={rgb, 255:red, 74; green, 144; blue, 226 }  ,draw opacity=1 ][line width=2.25]  (670.3,587.2) .. controls (670.3,584.33) and (672.63,582) .. (675.5,582) -- (908.8,582) .. controls (911.67,582) and (914,584.33) .. (914,587.2) -- (914,602.8) .. controls (914,605.67) and (911.67,608) .. (908.8,608) -- (675.5,608) .. controls (672.63,608) and (670.3,605.67) .. (670.3,602.8) -- cycle ;
\draw  [color={rgb, 255:red, 65; green, 117; blue, 5 }  ,draw opacity=1 ][line width=2.25]  (334,759.2) .. controls (334,756.33) and (336.33,754) .. (339.2,754) -- (574.8,754) .. controls (577.67,754) and (580,756.33) .. (580,759.2) -- (580,774.8) .. controls (580,777.67) and (577.67,780) .. (574.8,780) -- (339.2,780) .. controls (336.33,780) and (334,777.67) .. (334,774.8) -- cycle ;
\draw [color={rgb, 255:red, 189; green, 16; blue, 224 }  ,draw opacity=1 ][line width=1.5]    (597.5,305) .. controls (455.21,348.78) and (379.26,389.59) .. (278.03,511.16) ;
\draw [shift={(276.5,513)}, rotate = 309.66999999999996] [color={rgb, 255:red, 189; green, 16; blue, 224 }  ,draw opacity=1 ][line width=1.5]    (14.21,-4.28) .. controls (9.04,-1.82) and (4.3,-0.39) .. (0,0) .. controls (4.3,0.39) and (9.04,1.82) .. (14.21,4.28)   ;

\draw [color={rgb, 255:red, 155; green, 155; blue, 155 }  ,draw opacity=1 ][line width=1.5]    (282.5,520) .. controls (333.25,497.12) and (516.65,394.04) .. (602.22,317.16) ;
\draw [shift={(603.5,316)}, rotate = 497.83] [color={rgb, 255:red, 155; green, 155; blue, 155 }  ,draw opacity=1 ][line width=1.5]    (14.21,-4.28) .. controls (9.04,-1.82) and (4.3,-0.39) .. (0,0) .. controls (4.3,0.39) and (9.04,1.82) .. (14.21,4.28)   ;

\draw (144,597) node [scale=1.2] [align=left] {Speed : RampDown(130,138)};
\draw (128,419) node [scale=1.2] [align=left] {\textcolor[rgb]{0.74,0.06,0.88}{MaxSpeed : Interval}\textcolor[rgb]{0.74,0.06,0.88}{(}\textcolor[rgb]{0.74,0.06,0.88}{129,131}\textcolor[rgb]{0.74,0.06,0.88}{)}};
\draw (134,439) node [scale=1.2] [align=left] {\textcolor[rgb]{0.74,0.06,0.88}{Visibility : RampUp}\textcolor[rgb]{0.74,0.06,0.88}{(}\textcolor[rgb]{0.74,0.06,0.88}{0.045,0.05}\textcolor[rgb]{0.74,0.06,0.88}{)}};
\draw (148,458) node [scale=1.2] [align=left] {\textcolor[rgb]{0.74,0.06,0.88}{Precipitation : RampDown}\textcolor[rgb]{0.74,0.06,0.88}{(}\textcolor[rgb]{0.74,0.06,0.88}{0.0,0.5}\textcolor[rgb]{0.74,0.06,0.88}{)}};
\draw (245.5,541.5) node  [align=left] {130 km/h};
\draw (793,594) node [scale=1.2] [align=left] {Speed : RampDown(110,116)};
\draw (796,417) node [scale=1.2] [align=left] {\textcolor[rgb]{0.29,0.56,0.89}{MaxSpeed : Interval}\textcolor[rgb]{0.29,0.56,0.89}{(}\textcolor[rgb]{0.29,0.56,0.89}{129,131)}};
\draw (802,436) node [scale=1.2] [align=left] {\textcolor[rgb]{0.29,0.56,0.89}{Visibility : RampUp}\textcolor[rgb]{0.29,0.56,0.89}{(}\textcolor[rgb]{0.29,0.56,0.89}{0.045,0.05}\textcolor[rgb]{0.29,0.56,0.89}{)}};
\draw (804,454) node [scale=1.2] [align=left] {\textcolor[rgb]{0.29,0.56,0.89}{Precipitation : RampUp}\textcolor[rgb]{0.29,0.56,0.89}{(}\textcolor[rgb]{0.29,0.56,0.89}{0.5,1.0}\textcolor[rgb]{0.29,0.56,0.89}{)}};
\draw (679.5,539.5) node  [align=left] {130 km/h\\ \ RAINY};
\draw (459,767) node [scale=1.2] [align=left] {Speed <= RampDown(50,56)};
\draw (659,691) node [scale=1.2] [align=left] {\textcolor[rgb]{0.25,0.46,0.02}{Visibility : RampDown}\textcolor[rgb]{0.25,0.46,0.02}{(}\textcolor[rgb]{0.25,0.46,0.02}{0.045,0.05}\textcolor[rgb]{0.25,0.46,0.02}{)}};
\draw (457.5,705.5) node  [align=left] { \ \ \ \ LOW \\VISIBILITY};
\draw (249,301) node [scale=1.2] [align=left] {Speed : RampDown(110,116)};
\draw (327,174) node [scale=1.2] [align=left] {\textcolor[rgb]{0.31,0.89,0.76}{MaxSpeed : Interval}\textcolor[rgb]{0.31,0.89,0.76}{(}\textcolor[rgb]{0.31,0.89,0.76}{109,111}\textcolor[rgb]{0.31,0.89,0.76}{)}};
\draw (333,193) node [scale=1.2] [align=left] {\textcolor[rgb]{0.31,0.89,0.76}{Visibility : RampUp}\textcolor[rgb]{0.31,0.89,0.76}{(}\textcolor[rgb]{0.31,0.89,0.76}{0.045,0.05}\textcolor[rgb]{0.31,0.89,0.76}{)}};
\draw (347,212) node [scale=1.2] [align=left] {\textcolor[rgb]{0.31,0.89,0.76}{Precipitation : RampDown}\textcolor[rgb]{0.31,0.89,0.76}{(}\textcolor[rgb]{0.31,0.89,0.76}{0.0,0.5}\textcolor[rgb]{0.31,0.89,0.76}{)}};
\draw (416.5,298.5) node  [align=left] {110 km/h};
\draw (697,172) node [scale=1.2] [align=left] {\textcolor[rgb]{0.5,0.5,0.5}{MaxSpeed : Interval}\textcolor[rgb]{0.5,0.5,0.5}{(}\textcolor[rgb]{0.5,0.5,0.5}{109,111)}};
\draw (705,191) node [scale=1.2] [align=left] {\textcolor[rgb]{0.5,0.5,0.5}{Visibility : RampUp}\textcolor[rgb]{0.5,0.5,0.5}{(}\textcolor[rgb]{0.5,0.5,0.5}{0.045,0.05}\textcolor[rgb]{0.5,0.5,0.5}{)}};
\draw (707,209) node [scale=1.2] [align=left] {\textcolor[rgb]{0.5,0.5,0.5}{Precipitation : RampUp}\textcolor[rgb]{0.5,0.5,0.5}{(}\textcolor[rgb]{0.5,0.5,0.5}{0.5,1.0}\textcolor[rgb]{0.5,0.5,0.5}{)}};
\draw (638.5,292.5) node  [align=left] {110 km/h\\ \ RAINY};
\draw (806,294) node [scale=1.2] [align=left] {Speed : RampDown(100,106)};

\end{tikzpicture}

\caption{\label{partial} Ev-IOHMM model (partial) corresponding to the scenario 1. The input of the model is multivariate, defined with the maximum speed allowed at time $t$ (MaxSpeed whose values are computed as described in Table.\ref{speed_evhmm}), the current visibility (Visibility) and the precipitation (Precipitation). The output corresponds to the expected speed value (Speed) while being in each state. Some inputs can be inhibited depending on the state-transition considered. For instance, the state-transition leading state 'LOW VISIBILITY' to be reached only depends on the visibility value.}
\end{figure}
\medskip

\begin{figure}[ht!]
\centering
\tikzset{every picture/.style={line width=0.75pt}} 

\begin{tikzpicture}[scale=0.5, every node/.style={scale=0.5}, x=0.75pt,y=0.75pt,yscale=-1,xscale=1]

\draw  [color={rgb, 255:red, 189; green, 16; blue, 224 }  ,draw opacity=1 ][line width=2.25]  (263,437.5) .. controls (263,414.03) and (282.03,395) .. (305.5,395) .. controls (328.97,395) and (348,414.03) .. (348,437.5) .. controls (348,460.97) and (328.97,480) .. (305.5,480) .. controls (282.03,480) and (263,460.97) .. (263,437.5) -- cycle ;
\draw [color={rgb, 255:red, 189; green, 16; blue, 224 }  ,draw opacity=1 ][line width=2.25]    (280.5,402) .. controls (268.74,338.3) and (206.08,392.73) .. (261.93,421.29) ;
\draw [shift={(265.5,423)}, rotate = 204.3] [color={rgb, 255:red, 189; green, 16; blue, 224 }  ,draw opacity=1 ][line width=2.25]    (17.49,-5.26) .. controls (11.12,-2.23) and (5.29,-0.48) .. (0,0) .. controls (5.29,0.48) and (11.12,2.23) .. (17.49,5.26)   ;

\draw  [color={rgb, 255:red, 74; green, 144; blue, 226 }  ,draw opacity=1 ][line width=2.25]  (748,516.5) .. controls (748,493.03) and (767.03,474) .. (790.5,474) .. controls (813.97,474) and (833,493.03) .. (833,516.5) .. controls (833,539.97) and (813.97,559) .. (790.5,559) .. controls (767.03,559) and (748,539.97) .. (748,516.5) -- cycle ;
\draw [color={rgb, 255:red, 74; green, 144; blue, 226 }  ,draw opacity=1 ][line width=2.25]    (813.5,479) .. controls (814.49,415.96) and (895.03,469.35) .. (831.5,494.85) ;
\draw [shift={(828.5,496)}, rotate = 340.08000000000004] [color={rgb, 255:red, 74; green, 144; blue, 226 }  ,draw opacity=1 ][line width=2.25]    (17.49,-5.26) .. controls (11.12,-2.23) and (5.29,-0.48) .. (0,0) .. controls (5.29,0.48) and (11.12,2.23) .. (17.49,5.26)   ;

\draw [color={rgb, 255:red, 74; green, 144; blue, 226 }  ,draw opacity=1 ][line width=1.5]    (344.5,416) .. controls (531.56,427.94) and (585.96,439.88) .. (747.06,509.94) ;
\draw [shift={(749.5,511)}, rotate = 203.54] [color={rgb, 255:red, 74; green, 144; blue, 226 }  ,draw opacity=1 ][line width=1.5]    (14.21,-4.28) .. controls (9.04,-1.82) and (4.3,-0.39) .. (0,0) .. controls (4.3,0.39) and (9.04,1.82) .. (14.21,4.28)   ;

\draw [color={rgb, 255:red, 189; green, 16; blue, 224 }  ,draw opacity=1 ][line width=1.5]    (747.5,524) .. controls (561.43,509.08) and (452.59,467.42) .. (348.08,430.56) ;
\draw [shift={(346.5,430)}, rotate = 379.40999999999997] [color={rgb, 255:red, 189; green, 16; blue, 224 }  ,draw opacity=1 ][line width=1.5]    (14.21,-4.28) .. controls (9.04,-1.82) and (4.3,-0.39) .. (0,0) .. controls (4.3,0.39) and (9.04,1.82) .. (14.21,4.28)   ;

\draw  [color={rgb, 255:red, 65; green, 117; blue, 5 }  ,draw opacity=1 ][line width=2.25]  (526,803.5) .. controls (526,780.03) and (545.03,761) .. (568.5,761) .. controls (591.97,761) and (611,780.03) .. (611,803.5) .. controls (611,826.97) and (591.97,846) .. (568.5,846) .. controls (545.03,846) and (526,826.97) .. (526,803.5) -- cycle ;
\draw [color={rgb, 255:red, 65; green, 117; blue, 5 }  ,draw opacity=1 ][line width=2.25]    (611,803.5) .. controls (655.82,779.37) and (671.04,840.13) .. (606.5,829.53) ;
\draw [shift={(603.5,829)}, rotate = 370.82] [color={rgb, 255:red, 65; green, 117; blue, 5 }  ,draw opacity=1 ][line width=2.25]    (17.49,-5.26) .. controls (11.12,-2.23) and (5.29,-0.48) .. (0,0) .. controls (5.29,0.48) and (11.12,2.23) .. (17.49,5.26)   ;

\draw [color={rgb, 255:red, 65; green, 117; blue, 5 }  ,draw opacity=1 ][line width=1.5]    (771.5,556) .. controls (722.75,662.47) and (681.91,716.46) .. (609.59,786.94) ;
\draw [shift={(608.5,788)}, rotate = 315.8] [color={rgb, 255:red, 65; green, 117; blue, 5 }  ,draw opacity=1 ][line width=1.5]    (14.21,-4.28) .. controls (9.04,-1.82) and (4.3,-0.39) .. (0,0) .. controls (4.3,0.39) and (9.04,1.82) .. (14.21,4.28)   ;

\draw [color={rgb, 255:red, 74; green, 144; blue, 226 }  ,draw opacity=1 ][line width=1.5]    (595.5,771) .. controls (629.33,693.39) and (668.11,628.65) .. (760.11,550.18) ;
\draw [shift={(761.5,549)}, rotate = 499.65] [color={rgb, 255:red, 74; green, 144; blue, 226 }  ,draw opacity=1 ][line width=1.5]    (14.21,-4.28) .. controls (9.04,-1.82) and (4.3,-0.39) .. (0,0) .. controls (4.3,0.39) and (9.04,1.82) .. (14.21,4.28)   ;

\draw [color={rgb, 255:red, 65; green, 117; blue, 5 }  ,draw opacity=1 ][line width=1.5]    (342.5,462) .. controls (364.39,549.56) and (442.71,667.81) .. (532.15,776.36) ;
\draw [shift={(533.5,778)}, rotate = 230.45] [color={rgb, 255:red, 65; green, 117; blue, 5 }  ,draw opacity=1 ][line width=1.5]    (14.21,-4.28) .. controls (9.04,-1.82) and (4.3,-0.39) .. (0,0) .. controls (4.3,0.39) and (9.04,1.82) .. (14.21,4.28)   ;

\draw [color={rgb, 255:red, 189; green, 16; blue, 224 }  ,draw opacity=1 ][line width=1.5]    (540.5,771) .. controls (510.65,680.46) and (445.16,552.29) .. (346.98,451.52) ;
\draw [shift={(345.5,450)}, rotate = 405.57] [color={rgb, 255:red, 189; green, 16; blue, 224 }  ,draw opacity=1 ][line width=1.5]    (14.21,-4.28) .. controls (9.04,-1.82) and (4.3,-0.39) .. (0,0) .. controls (4.3,0.39) and (9.04,1.82) .. (14.21,4.28)   ;

\draw  [color={rgb, 255:red, 80; green, 227; blue, 194 }  ,draw opacity=1 ][line width=2.25]  (485,275.5) .. controls (485,252.03) and (504.03,233) .. (527.5,233) .. controls (550.97,233) and (570,252.03) .. (570,275.5) .. controls (570,298.97) and (550.97,318) .. (527.5,318) .. controls (504.03,318) and (485,298.97) .. (485,275.5) -- cycle ;
\draw [color={rgb, 255:red, 80; green, 227; blue, 194 }  ,draw opacity=1 ][line width=2.25]    (502.5,240) .. controls (490.74,176.3) and (428.08,230.73) .. (483.93,259.29) ;
\draw [shift={(487.5,261)}, rotate = 204.3] [color={rgb, 255:red, 80; green, 227; blue, 194 }  ,draw opacity=1 ][line width=2.25]    (17.49,-5.26) .. controls (11.12,-2.23) and (5.29,-0.48) .. (0,0) .. controls (5.29,0.48) and (11.12,2.23) .. (17.49,5.26)   ;

\draw [color={rgb, 255:red, 189; green, 16; blue, 224 }  ,draw opacity=1 ][line width=1.5]    (496.5,304) .. controls (447.99,345.58) and (401.44,371.48) .. (328.71,399.16) ;
\draw [shift={(326.5,400)}, rotate = 339.27] [color={rgb, 255:red, 189; green, 16; blue, 224 }  ,draw opacity=1 ][line width=1.5]    (14.21,-4.28) .. controls (9.04,-1.82) and (4.3,-0.39) .. (0,0) .. controls (4.3,0.39) and (9.04,1.82) .. (14.21,4.28)   ;

\draw [color={rgb, 255:red, 80; green, 227; blue, 194 }  ,draw opacity=1 ][line width=1.5]    (487.15,300.21) .. controls (414.24,326.43) and (372.6,340.84) .. (313.5,396) ;

\draw [shift={(490.5,299)}, rotate = 160.2] [color={rgb, 255:red, 80; green, 227; blue, 194 }  ,draw opacity=1 ][line width=1.5]    (14.21,-4.28) .. controls (9.04,-1.82) and (4.3,-0.39) .. (0,0) .. controls (4.3,0.39) and (9.04,1.82) .. (14.21,4.28)   ;
\draw [color={rgb, 255:red, 65; green, 117; blue, 5 }  ,draw opacity=1 ][line width=1.5]    (534.5,317) .. controls (591.22,452.32) and (606.35,599.52) .. (565.13,760.57) ;
\draw [shift={(564.5,763)}, rotate = 284.53] [color={rgb, 255:red, 65; green, 117; blue, 5 }  ,draw opacity=1 ][line width=1.5]    (14.21,-4.28) .. controls (9.04,-1.82) and (4.3,-0.39) .. (0,0) .. controls (4.3,0.39) and (9.04,1.82) .. (14.21,4.28)   ;

\draw [color={rgb, 255:red, 80; green, 227; blue, 194 }  ,draw opacity=1 ][line width=1.5]    (551.5,762) .. controls (501.75,646.58) and (495.56,442.06) .. (524.07,320.82) ;
\draw [shift={(524.5,319)}, rotate = 463.48] [color={rgb, 255:red, 80; green, 227; blue, 194 }  ,draw opacity=1 ][line width=1.5]    (14.21,-4.28) .. controls (9.04,-1.82) and (4.3,-0.39) .. (0,0) .. controls (4.3,0.39) and (9.04,1.82) .. (14.21,4.28)   ;

\draw  [color={rgb, 255:red, 128; green, 128; blue, 128 }  ,draw opacity=1 ][line width=2.25]  (707,269.5) .. controls (707,246.03) and (726.03,227) .. (749.5,227) .. controls (772.97,227) and (792,246.03) .. (792,269.5) .. controls (792,292.97) and (772.97,312) .. (749.5,312) .. controls (726.03,312) and (707,292.97) .. (707,269.5) -- cycle ;
\draw [color={rgb, 255:red, 155; green, 155; blue, 155 }  ,draw opacity=1 ][line width=2.25]    (772.5,232) .. controls (773.49,168.96) and (854.03,222.35) .. (790.5,247.85) ;
\draw [shift={(787.5,249)}, rotate = 340.08000000000004] [color={rgb, 255:red, 155; green, 155; blue, 155 }  ,draw opacity=1 ][line width=2.25]    (17.49,-5.26) .. controls (11.12,-2.23) and (5.29,-0.48) .. (0,0) .. controls (5.29,0.48) and (11.12,2.23) .. (17.49,5.26)   ;

\draw [color={rgb, 255:red, 155; green, 155; blue, 155 }  ,draw opacity=1 ][line width=1.5]    (767.5,480) .. controls (740.77,419.61) and (735.6,391.56) .. (739.38,313.38) ;
\draw [shift={(739.5,311)}, rotate = 452.86] [color={rgb, 255:red, 155; green, 155; blue, 155 }  ,draw opacity=1 ][line width=1.5]    (14.21,-4.28) .. controls (9.04,-1.82) and (4.3,-0.39) .. (0,0) .. controls (4.3,0.39) and (9.04,1.82) .. (14.21,4.28)   ;

\draw [color={rgb, 255:red, 74; green, 144; blue, 226 }  ,draw opacity=1 ][line width=1.5]    (754.5,313) .. controls (771.33,342.7) and (779.34,400.82) .. (780.47,472.81) ;
\draw [shift={(780.5,475)}, rotate = 269.22] [color={rgb, 255:red, 74; green, 144; blue, 226 }  ,draw opacity=1 ][line width=1.5]    (14.21,-4.28) .. controls (9.04,-1.82) and (4.3,-0.39) .. (0,0) .. controls (4.3,0.39) and (9.04,1.82) .. (14.21,4.28)   ;

\draw [color={rgb, 255:red, 155; green, 155; blue, 155 }  ,draw opacity=1 ][line width=1.5]    (579.5,762) .. controls (608.36,572.95) and (650.08,435.38) .. (719.45,303.98) ;
\draw [shift={(720.5,302)}, rotate = 477.94] [color={rgb, 255:red, 155; green, 155; blue, 155 }  ,draw opacity=1 ][line width=1.5]    (14.21,-4.28) .. controls (9.04,-1.82) and (4.3,-0.39) .. (0,0) .. controls (4.3,0.39) and (9.04,1.82) .. (14.21,4.28)   ;

\draw [color={rgb, 255:red, 65; green, 117; blue, 5 }  ,draw opacity=1 ][line width=1.5]    (726.5,306) .. controls (702.62,434.36) and (651.02,614.19) .. (586.47,762.76) ;
\draw [shift={(585.5,765)}, rotate = 293.57] [color={rgb, 255:red, 65; green, 117; blue, 5 }  ,draw opacity=1 ][line width=1.5]    (14.21,-4.28) .. controls (9.04,-1.82) and (4.3,-0.39) .. (0,0) .. controls (4.3,0.39) and (9.04,1.82) .. (14.21,4.28)   ;

\draw [color={rgb, 255:red, 155; green, 155; blue, 155 }  ,draw opacity=1 ][line width=1.5]    (563.5,251) .. controls (610.78,232.29) and (660,228.12) .. (712.12,244.25) ;
\draw [shift={(714.5,245)}, rotate = 197.78] [color={rgb, 255:red, 155; green, 155; blue, 155 }  ,draw opacity=1 ][line width=1.5]    (14.21,-4.28) .. controls (9.04,-1.82) and (4.3,-0.39) .. (0,0) .. controls (4.3,0.39) and (9.04,1.82) .. (14.21,4.28)   ;

\draw [color={rgb, 255:red, 80; green, 227; blue, 194 }  ,draw opacity=1 ][line width=1.5]    (708.5,254) .. controls (649.89,266.31) and (627.18,265.04) .. (569.65,261.66) ;
\draw [shift={(567,261.5)}, rotate = 363.37] [color={rgb, 255:red, 80; green, 227; blue, 194 }  ,draw opacity=1 ][line width=1.5]    (14.21,-4.28) .. controls (9.04,-1.82) and (4.3,-0.39) .. (0,0) .. controls (4.3,0.39) and (9.04,1.82) .. (14.21,4.28)   ;

\draw [color={rgb, 255:red, 74; green, 144; blue, 226 }  ,draw opacity=1 ][line width=1.5]    (551.5,312) .. controls (568.42,341.85) and (628.89,432.09) .. (750.66,495.05) ;
\draw [shift={(752.5,496)}, rotate = 207.12] [color={rgb, 255:red, 74; green, 144; blue, 226 }  ,draw opacity=1 ][line width=1.5]    (14.21,-4.28) .. controls (9.04,-1.82) and (4.3,-0.39) .. (0,0) .. controls (4.3,0.39) and (9.04,1.82) .. (14.21,4.28)   ;

\draw [color={rgb, 255:red, 80; green, 227; blue, 194 }  ,draw opacity=1 ][line width=1.5]    (758.5,486) .. controls (709.99,396.9) and (643.84,343.08) .. (562.47,303.69) ;
\draw [shift={(560,302.5)}, rotate = 385.58000000000004] [color={rgb, 255:red, 80; green, 227; blue, 194 }  ,draw opacity=1 ][line width=1.5]    (14.21,-4.28) .. controls (9.04,-1.82) and (4.3,-0.39) .. (0,0) .. controls (4.3,0.39) and (9.04,1.82) .. (14.21,4.28)   ;

\draw  [color={rgb, 255:red, 189; green, 16; blue, 224 }  ,draw opacity=1 ][line width=2.25]  (12.5,486.2) .. controls (12.5,483.33) and (14.83,481) .. (17.7,481) -- (299.3,481) .. controls (302.17,481) and (304.5,483.33) .. (304.5,486.2) -- (304.5,501.8) .. controls (304.5,504.67) and (302.17,507) .. (299.3,507) -- (17.7,507) .. controls (14.83,507) and (12.5,504.67) .. (12.5,501.8) -- cycle ;
\draw  [color={rgb, 255:red, 80; green, 227; blue, 194 }  ,draw opacity=1 ][line width=2.25]  (189.5,270.2) .. controls (189.5,267.33) and (191.83,265) .. (194.7,265) -- (479.3,265) .. controls (482.17,265) and (484.5,267.33) .. (484.5,270.2) -- (484.5,285.8) .. controls (484.5,288.67) and (482.17,291) .. (479.3,291) -- (194.7,291) .. controls (191.83,291) and (189.5,288.67) .. (189.5,285.8) -- cycle ;
\draw  [color={rgb, 255:red, 128; green, 128; blue, 128 }  ,draw opacity=1 ][line width=2.25]  (792,263.2) .. controls (792,260.33) and (794.33,258) .. (797.2,258) -- (1082.3,258) .. controls (1085.17,258) and (1087.5,260.33) .. (1087.5,263.2) -- (1087.5,278.8) .. controls (1087.5,281.67) and (1085.17,284) .. (1082.3,284) -- (797.2,284) .. controls (794.33,284) and (792,281.67) .. (792,278.8) -- cycle ;
\draw  [color={rgb, 255:red, 74; green, 144; blue, 226 }  ,draw opacity=1 ][line width=2.25]  (782.3,565.2) .. controls (782.3,562.33) and (784.63,560) .. (787.5,560) -- (1071.3,560) .. controls (1074.17,560) and (1076.5,562.33) .. (1076.5,565.2) -- (1076.5,580.8) .. controls (1076.5,583.67) and (1074.17,586) .. (1071.3,586) -- (787.5,586) .. controls (784.63,586) and (782.3,583.67) .. (782.3,580.8) -- cycle ;
\draw  [color={rgb, 255:red, 65; green, 117; blue, 5 }  ,draw opacity=1 ][line width=2.25]  (418.5,852.2) .. controls (418.5,849.33) and (420.83,847) .. (423.7,847) -- (711.3,847) .. controls (714.17,847) and (716.5,849.33) .. (716.5,852.2) -- (716.5,867.8) .. controls (716.5,870.67) and (714.17,873) .. (711.3,873) -- (423.7,873) .. controls (420.83,873) and (418.5,870.67) .. (418.5,867.8) -- cycle ;
\draw  [color={rgb, 255:red, 245; green, 166; blue, 35 }  ,draw opacity=1 ][line width=2.25]  (317,672.5) .. controls (317,649.03) and (336.03,630) .. (359.5,630) .. controls (382.97,630) and (402,649.03) .. (402,672.5) .. controls (402,695.97) and (382.97,715) .. (359.5,715) .. controls (336.03,715) and (317,695.97) .. (317,672.5) -- cycle ;
\draw [color={rgb, 255:red, 245; green, 166; blue, 35 }  ,draw opacity=1 ][line width=2.25]    (334.5,637) .. controls (322.74,573.3) and (260.08,627.73) .. (315.93,656.29) ;
\draw [shift={(319.5,658)}, rotate = 204.3] [color={rgb, 255:red, 245; green, 166; blue, 35 }  ,draw opacity=1 ][line width=2.25]    (17.49,-5.26) .. controls (11.12,-2.23) and (5.29,-0.48) .. (0,0) .. controls (5.29,0.48) and (11.12,2.23) .. (17.49,5.26)   ;

\draw  [color={rgb, 255:red, 245; green, 166; blue, 35 }  ,draw opacity=1 ][line width=2.25]  (63.5,721.2) .. controls (63.5,718.33) and (65.83,716) .. (68.7,716) -- (353.3,716) .. controls (356.17,716) and (358.5,718.33) .. (358.5,721.2) -- (358.5,736.8) .. controls (358.5,739.67) and (356.17,742) .. (353.3,742) -- (68.7,742) .. controls (65.83,742) and (63.5,739.67) .. (63.5,736.8) -- cycle ;
\draw [color={rgb, 255:red, 189; green, 16; blue, 224 }  ,draw opacity=1 ][line width=1.5]    (349.5,631) .. controls (335.71,579.78) and (327.74,529.53) .. (315.08,481.21) ;
\draw [shift={(314.5,479)}, rotate = 435.14] [color={rgb, 255:red, 189; green, 16; blue, 224 }  ,draw opacity=1 ][line width=1.5]    (14.21,-4.28) .. controls (9.04,-1.82) and (4.3,-0.39) .. (0,0) .. controls (4.3,0.39) and (9.04,1.82) .. (14.21,4.28)   ;

\draw [color={rgb, 255:red, 245; green, 166; blue, 35 }  ,draw opacity=1 ][line width=1.5]    (365.67,626.88) .. controls (352.18,575.76) and (344.24,521.02) .. (331.5,473) ;

\draw [shift={(366.5,630)}, rotate = 254.93] [color={rgb, 255:red, 245; green, 166; blue, 35 }  ,draw opacity=1 ][line width=1.5]    (14.21,-4.28) .. controls (9.04,-1.82) and (4.3,-0.39) .. (0,0) .. controls (4.3,0.39) and (9.04,1.82) .. (14.21,4.28)   ;
\draw [color={rgb, 255:red, 80; green, 227; blue, 194 }  ,draw opacity=1 ][line width=1.5]    (377.5,634) .. controls (397.4,544.45) and (418.29,452.92) .. (500.26,309.17) ;
\draw [shift={(501.5,307)}, rotate = 479.79] [color={rgb, 255:red, 80; green, 227; blue, 194 }  ,draw opacity=1 ][line width=1.5]    (14.21,-4.28) .. controls (9.04,-1.82) and (4.3,-0.39) .. (0,0) .. controls (4.3,0.39) and (9.04,1.82) .. (14.21,4.28)   ;

\draw [color={rgb, 255:red, 245; green, 166; blue, 35 }  ,draw opacity=1 ][line width=1.5]    (392.17,637.89) .. controls (447.16,535.55) and (479.85,463.49) .. (514.5,314) ;

\draw [shift={(390.5,641)}, rotate = 298.3] [color={rgb, 255:red, 245; green, 166; blue, 35 }  ,draw opacity=1 ][line width=1.5]    (14.21,-4.28) .. controls (9.04,-1.82) and (4.3,-0.39) .. (0,0) .. controls (4.3,0.39) and (9.04,1.82) .. (14.21,4.28)   ;
\draw [color={rgb, 255:red, 128; green, 128; blue, 128 }  ,draw opacity=1 ][line width=1.5]    (396.5,648) .. controls (460.18,537.55) and (555.54,405.33) .. (707.21,287.77) ;
\draw [shift={(709.5,286)}, rotate = 502.36] [color={rgb, 255:red, 128; green, 128; blue, 128 }  ,draw opacity=1 ][line width=1.5]    (14.21,-4.28) .. controls (9.04,-1.82) and (4.3,-0.39) .. (0,0) .. controls (4.3,0.39) and (9.04,1.82) .. (14.21,4.28)   ;

\draw [color={rgb, 255:red, 245; green, 166; blue, 35 }  ,draw opacity=1 ][line width=1.5]    (405.03,659.99) .. controls (521.44,528.14) and (617.49,435.58) .. (715.5,295) ;

\draw [shift={(401.5,664)}, rotate = 311.37] [color={rgb, 255:red, 245; green, 166; blue, 35 }  ,draw opacity=1 ][line width=1.5]    (14.21,-4.28) .. controls (9.04,-1.82) and (4.3,-0.39) .. (0,0) .. controls (4.3,0.39) and (9.04,1.82) .. (14.21,4.28)   ;
\draw [color={rgb, 255:red, 245; green, 166; blue, 35 }  ,draw opacity=1 ][line width=1.5]    (405.81,673.1) .. controls (530.87,594.58) and (599.03,573.57) .. (750.5,531) ;

\draw [shift={(402,675.5)}, rotate = 327.73] [color={rgb, 255:red, 245; green, 166; blue, 35 }  ,draw opacity=1 ][line width=1.5]    (14.21,-4.28) .. controls (9.04,-1.82) and (4.3,-0.39) .. (0,0) .. controls (4.3,0.39) and (9.04,1.82) .. (14.21,4.28)   ;
\draw [color={rgb, 255:red, 74; green, 144; blue, 226 }  ,draw opacity=1 ][line width=1.5]    (399.5,689) .. controls (511.94,657.16) and (635.26,613.44) .. (753.71,540.11) ;
\draw [shift={(755.5,539)}, rotate = 508.12] [color={rgb, 255:red, 74; green, 144; blue, 226 }  ,draw opacity=1 ][line width=1.5]    (14.21,-4.28) .. controls (9.04,-1.82) and (4.3,-0.39) .. (0,0) .. controls (4.3,0.39) and (9.04,1.82) .. (14.21,4.28)   ;

\draw [color={rgb, 255:red, 245; green, 166; blue, 35 }  ,draw opacity=1 ][line width=1.5]    (395.75,701.61) .. controls (466.13,736.58) and (487.09,755.65) .. (525.5,798) ;

\draw [shift={(392.5,700)}, rotate = 26.25] [color={rgb, 255:red, 245; green, 166; blue, 35 }  ,draw opacity=1 ][line width=1.5]    (14.21,-4.28) .. controls (9.04,-1.82) and (4.3,-0.39) .. (0,0) .. controls (4.3,0.39) and (9.04,1.82) .. (14.21,4.28)   ;
\draw [color={rgb, 255:red, 65; green, 117; blue, 5 }  ,draw opacity=1 ][line width=1.5]    (520.91,809.66) .. controls (462.59,787.56) and (415.84,752.14) .. (383.5,710) ;

\draw [shift={(524.5,811)}, rotate = 200.14] [color={rgb, 255:red, 65; green, 117; blue, 5 }  ,draw opacity=1 ][line width=1.5]    (14.21,-4.28) .. controls (9.04,-1.82) and (4.3,-0.39) .. (0,0) .. controls (4.3,0.39) and (9.04,1.82) .. (14.21,4.28)   ;
\draw [color={rgb, 255:red, 189; green, 16; blue, 224 }  ,draw opacity=1 ][line width=1.5]    (708.5,278) .. controls (569.2,368.55) and (477.42,396.72) .. (341.55,410.79) ;
\draw [shift={(339.5,411)}, rotate = 354.16999999999996] [color={rgb, 255:red, 189; green, 16; blue, 224 }  ,draw opacity=1 ][line width=1.5]    (14.21,-4.28) .. controls (9.04,-1.82) and (4.3,-0.39) .. (0,0) .. controls (4.3,0.39) and (9.04,1.82) .. (14.21,4.28)   ;

\draw [color={rgb, 255:red, 155; green, 155; blue, 155 }  ,draw opacity=1 ][line width=1.5]    (334.5,406) .. controls (382.26,387.1) and (604.26,296.91) .. (704.99,270.39) ;
\draw [shift={(706.5,270)}, rotate = 525.4300000000001] [color={rgb, 255:red, 155; green, 155; blue, 155 }  ,draw opacity=1 ][line width=1.5]    (14.21,-4.28) .. controls (9.04,-1.82) and (4.3,-0.39) .. (0,0) .. controls (4.3,0.39) and (9.04,1.82) .. (14.21,4.28)   ;

\draw (305.5,437.5) node  [align=left] {GIVE\\ WAY};
\draw (898,431) node [scale=1.2] [align=left] {\textcolor[rgb]{0.29,0.56,0.89}{Stop : Interval}\textcolor[rgb]{0.29,0.56,0.89}{(}\textcolor[rgb]{0.29,0.56,0.89}{0.9,1.1}\textcolor[rgb]{0.29,0.56,0.89}{)}};
\draw (790.5,516.5) node  [align=left] {STOP};
\draw (568.5,798.5) node  [align=left] { \ \ \ \ NO\\FEATURE};
\draw (339,277) node [scale=1.2] [align=left] {Acceleration : Trapezoidal(-3,-1,1,3)};
\draw (454,189) node [scale=1.2] [align=left] {\textcolor[rgb]{0.31,0.89,0.76}{Hump : Interval}\textcolor[rgb]{0.31,0.89,0.76}{(}\textcolor[rgb]{0.31,0.89,0.76}{0.9,1.1}\textcolor[rgb]{0.31,0.89,0.76}{)}};
\draw (527.5,275.5) node  [align=left] {HUMP};
\draw (749.5,269.5) node  [align=left] {CHOKER};
\draw (793,186) node [scale=1.2] [align=left] {\textcolor[rgb]{0.5,0.5,0.5}{Choker : Interval}\textcolor[rgb]{0.5,0.5,0.5}{(}\textcolor[rgb]{0.5,0.5,0.5}{0.9,1.1}\textcolor[rgb]{0.5,0.5,0.5}{)}};
\draw (941,270) node [scale=1.2] [align=left] {Acceleration : Trapezoidal(-3,-1,1,3)};
\draw (159,493) node [scale=1.2] [align=left] {Acceleration : Trapezoidal(-3,-1,1,3)};
\draw (235,353) node [scale=1.2] [align=left] {\textcolor[rgb]{0.74,0.06,0.88}{Give\_way : Interval}\textcolor[rgb]{0.74,0.06,0.88}{(}\textcolor[rgb]{0.74,0.06,0.88}{0.9,1.1}\textcolor[rgb]{0.74,0.06,0.88}{)}};
\draw (930,572) node [scale=1.2] [align=left] {Acceleration : Trapezoidal(-3,-1,1,3)};
\draw (568,860) node [scale=1.2] [align=left] {Acceleration : Trapezoidal(-7,-1,1,7)};
\draw (777,761) node [scale=1.2,color={rgb, 255:red, 65; green, 117; blue, 5 }  ,opacity=1 ] [align=left] {\textcolor[rgb]{0.25,0.46,0.02}{Hump : RampDown}\textcolor[rgb]{0.25,0.46,0.02}{(}\textcolor[rgb]{0.25,0.46,0.02}{0.0,0.5}\textcolor[rgb]{0.25,0.46,0.02}{)}};
\draw (792,783) node [scale=1.2,color={rgb, 255:red, 65; green, 117; blue, 5 }  ,opacity=1 ] [align=left] {\textcolor[rgb]{0.25,0.46,0.02}{Give\_way : RampDown}\textcolor[rgb]{0.25,0.46,0.02}{(}\textcolor[rgb]{0.25,0.46,0.02}{0.0,0.5}\textcolor[rgb]{0.25,0.46,0.02}{)}};
\draw (772,804) node [scale=1.2,color={rgb, 255:red, 65; green, 117; blue, 5 }  ,opacity=1 ] [align=left] {\textcolor[rgb]{0.25,0.46,0.02}{Stop : RampDown}\textcolor[rgb]{0.25,0.46,0.02}{(}\textcolor[rgb]{0.25,0.46,0.02}{0.0,0.5}\textcolor[rgb]{0.25,0.46,0.02}{)}};
\draw (782,825) node [scale=1.2,color={rgb, 255:red, 65; green, 117; blue, 5 }  ,opacity=1 ] [align=left] {\textcolor[rgb]{0.25,0.46,0.02}{Choker : RampDown}\textcolor[rgb]{0.25,0.46,0.02}{(}\textcolor[rgb]{0.25,0.46,0.02}{0.0,0.5}\textcolor[rgb]{0.25,0.46,0.02}{)}};
\draw (213,727) node [scale=1.2] [align=left] {Acceleration : Trapezoidal(-3,-1,1,3)};
\draw (216,589) node [scale=1.2] [align=left] {\textcolor[rgb]{0.96,0.65,0.14}{Roundabout : Interval}\textcolor[rgb]{0.96,0.65,0.14}{(}\textcolor[rgb]{0.96,0.65,0.14}{0.9,1.1}\textcolor[rgb]{0.96,0.65,0.14}{)}};
\draw (359.5,672.5) node  [align=left] {ROUND\\ABOUT};
\end{tikzpicture}

\caption{\label{accelmodel} Ev-IOHMM corresponding to scenario 2. The acceleration of the vehicle (output of the model) is constrained depending on specific features (inputs of the model). }
\end{figure}
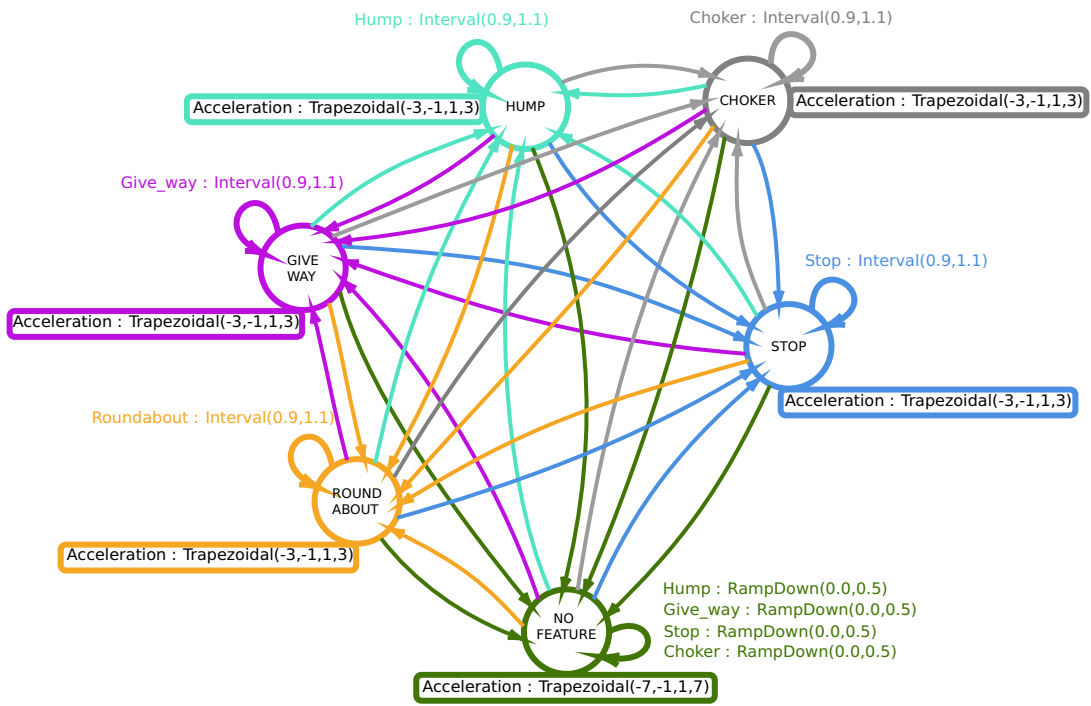

\noindent The model for the second scenario is built upon the same approach. The inputs of the associated Ev-IOHMM are gathered from some OSM features of interest, i.e, traffic\_calming:hump,  traffic\_calming:choker, highway:stop and highway:give\_way, taking value of "1.0" if the feature is detected around, "0.0" otherwise. The output of the Ev-IOHMM corresponds to the constraint on the acceleration and deceleration while being in each state. Instant acceleration and deceleration of the vehicle is computed based on the delta of speed and the delta of the distance travelled between two time steps ($accel = \frac{v_{(t+n)}-v_{(t)}}{n}$). The corresponding Ev-HMM model is depicted in Fig.\ref{accelmodel}. 

\begin{note}
\noindent In these scenarios, we do assume the behavioral constraints are defined by the designer of the model. For instance, for the first scenario, the maximum speed limits in force are subject to tolerances inherent to radar systems accuracy (generally, in the 5\% range) that cannot be retrieved from learning. For the second scenario, tolerances may be adjusted based on users' feedback or from their preferences. 
\end{note}

\subsection{Results}

\begin{figure}[ht!]
  \centering
  \begin{subfigure}[b]{0.537\linewidth}
    \includegraphics[width=\linewidth]{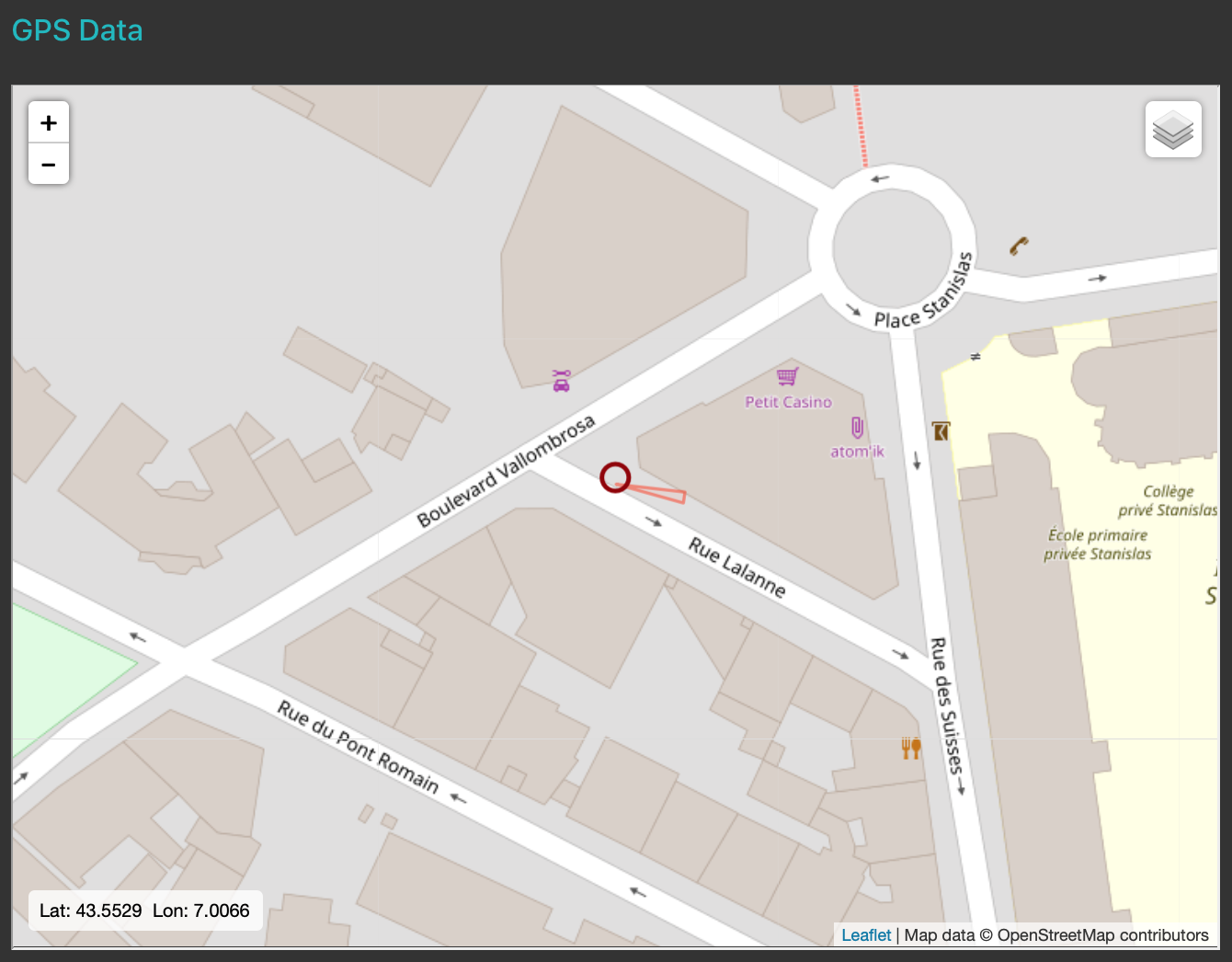}
  \end{subfigure}
  \begin{subfigure}[b]{0.4\linewidth}
    \includegraphics[width=\linewidth]{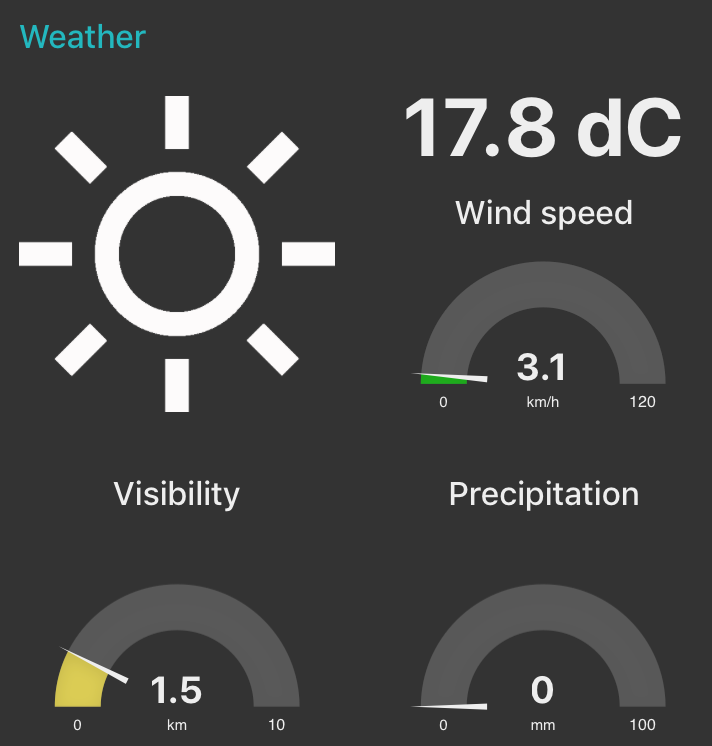}
  \end{subfigure}
  \caption{Vehicle position and weather information visualization.}
  \label{nr1}
\end{figure}

\begin{figure}[ht!]
  \centering
  \includegraphics[width=0.7\linewidth]{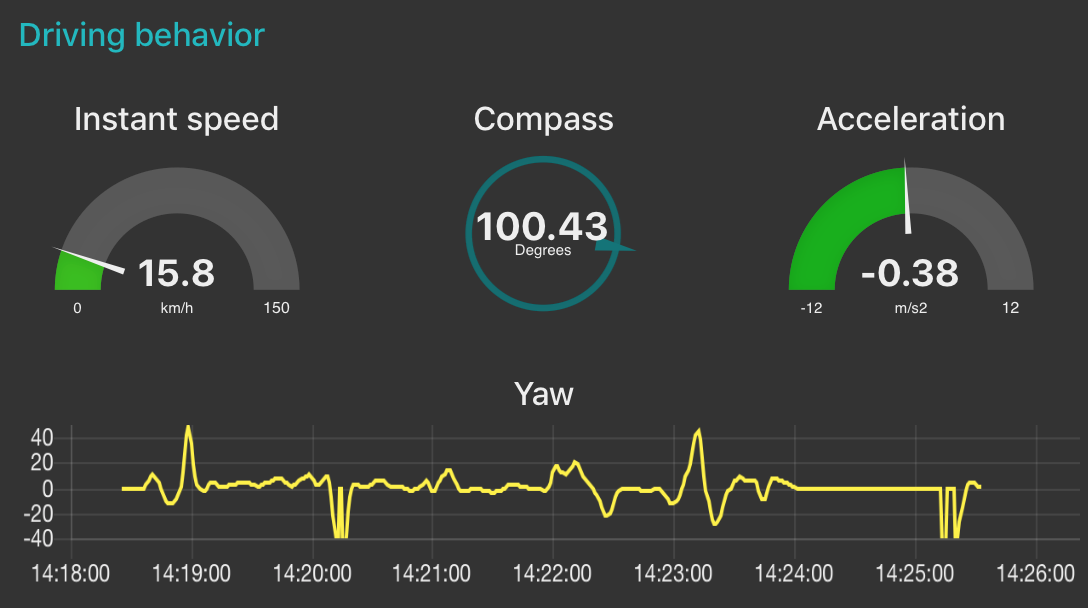}
  \caption{Driving behavior information consists in monitoring the current speed of the vehicle along with the instant acceleration and orientation. Although not used in the proposed scenario, the orientation (Yaw) may be used against OSM restrictions such as "no left turn", "no right turn", etc.}
  \label{nr2}
\end{figure}

\noindent To complete the experimentations, a C\# library has been developed based on the Matlab functions developed by Philippe Smets \cite{smets}. The library has been further extended with combination rules (CRC and DRC \cite{smets2008belief}) and normalization rules (Dempster \cite{yager1987dempster}, Yager \cite{yager1996normalization}, and Dubois-Prade \cite{martin2008general}). A mobile phone application is further used for recording GPS data from several drivers. The traces recorded are then post-processed in order to add, based on latitude and longitude information, the OSM infrastructure features of interest near the location of the vehicle along with the weather data, based on timestamp. The post processor aims at generating a dataset that can be replayed from a graphical interface built using Node-Red \cite{nodered} and in which the Ev-IOHMM models are instantiated for evaluation (see Fig.\ref{nr1} and Fig.\ref{nr2}). The degree of effectiveness is then assessed using the Ev-IOHMM models described in Fig.\ref{partial} and Fig.\ref{accelmodel}.  

\begin{figure}[ht!]
  \centering
  \includegraphics[width=0.99\linewidth]{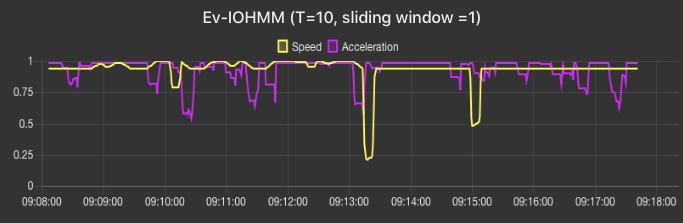}
  \caption{Assessment of the degree of effectiveness from Ev-IOHMM models described in Fig.\ref{partial} and Fig.\ref{accelmodel} where the sequence length is equal to 10s (observations are sampled each second). A sliding window is applied : after the first observation sequence is gathered, each new observation triggers the computation of the degree of effectiveness (sliding window = 1).}
  \label{nr4}
\end{figure}

\noindent Results are provided in Fig.\ref{nr4}. It is worth noting that the proposed approach is not predictive, i.e. the degree of effectiveness is computed based on the last $T$ observations. The length $T$ depends on the sampling rate of the sensors, the Ev-IOHMM model and the physical process  whose evolution is observed. By applying a sliding window on the observations, once the window has been filled up with $T$ observations, one can get the computation of the degree of effectiveness performed upon each new observation. Results obtained may help designers of autonomous vehicles to benchmark the controllers for different situations. 
\color{black}
\section{Conclusion and perspectives}
\noindent Cyber-Physical Systems (CPS) are computing systems whose purposes are achieved from interactions with the physical world by means of transducers (sensors and actuators). These systems pose new challenges in terms of dependability, the evolution of the physical systems they control being affected by unanticipated physical processes over which they have no control and which may potentially hamper the achievement of their purposes. It is now recognized that designers of such systems can no longer lean, at design time, on comprehensive and reliable models for anticipating and removing faults that may arise once these systems are deployed. Instead, they have to be monitored \textit{in vivo} and quantitatively evaluated for effectiveness throughout their life cycle.
\medskip

\noindent In this paper, we formally defined and generalized the notion of effectiveness in the context of the measure and viability theories. We further detailed the mathematical properties the measure has to comply with in the context of assessing CPS effectiveness. The measure is further developed in the Transferable Belief Model (TBM) framework, an elaboration of the Dempster-Shafer Theory (DST) of evidence. The proposed approach is intended to have several applications in the context of benchmarking and assessment of Quality of Experience (QoE)\cite{mitra2014qoe}:
\begin{itemize}
	\item[--] The measure can be used as a benchmarking tool. For instance, let us consider the case of autonomous driving. One might be interested in comparing algorithms used to control autonomous vehicles according to the highway code. As a future work, we do plan to apply the proposed approach to this use-case based on the UAH-DriveSet \cite{romera2016need}. 
	\item[--] Assuming the expected evolution of the physical system being driven by the CPS is specified by the users (e.g. through end-user programming \cite{coutaz2014early}), the degree of effectiveness might give a direct insight on users' satisfaction, i.e. QoE as an assessment of the human satisfaction when interacting with technology and business entities in a particular context.  
\end{itemize} 
\noindent The proposed approach may also provide an added value in self-adaptive systems:
\begin{itemize}
	\item[--] In the context of the Internet of Things (IoT), many physical devices now expose services available to ubiquitous computing systems leveraging them for composing the so-called ambient applications (e.g. smart-home, smart-city, etc.). The question then arises for these systems of how to select the relevant services. Current approaches rely on semantic annotations used to formally describe the services \cite{rocher2015run}\cite{kim2017smart}. While this approach is relevant, (1) semantic annotations are pure models, agnostic to the target operational environment, (2) the behavior of a composed application cannot be inferred solely given the individual behavior of the services it is composed with. Hence, observing the concrete behavior of these systems and providing them with a feedback through the measure of effectiveness would help them selecting more appropriate services over time. 
	\item[--] More generally, self-adaptive systems pose new challenges in term of assurance, i.e. the ability to provide evidence that these systems satisfy their behavioral requirements, irrespective of the adaptations over time \cite{de2014software}. One may envision leveraging the assessment proposed in this paper within a feedback loop providing CPS with \textit{self-awareness} capability allowing them to react towards any deviation.    
\end{itemize}
\noindent However, although promising, the proposed approach suffers from limitations :
\begin{itemize}
    \item[--] The approach is not predictive, the measure is computed based on past events. This could be a problem for safety critical CPS for which an immediate response is required,
    \item[--] As such, the Ev-IOHMM cannot manage temporal constraints which are of importance in the CPS context. Following what has been done on Hidden Semi-Markov Models (HSMM)\cite{yu2010hidden}, we do plan to develop the Ev-IOHSMM where temporal constraints are used to specify the maximum amount of time allowed to switch from one state to another one or to specify the maximum time one is allowed to stay in a particular state,
    \item[--] The Ev-IOHMM is memory greedy, it implies $2^{|\Omega|}\times2^{|\Omega|}$ elements in the state-transition matrix $A$. For instance, for the first scenario described in section \ref{app}, the model of the expected behavior contains 11 states leading a 2048$\times$2048 state-transition matrix. The complexity in time of the forward algorithm described in \ref{evfa}  is $\mathcal{O}(2^{2|\Omega|}T)$ when using commonalities $q^{\Omega}$. By using plausibilities $pl^{\Omega}$, the complexity is $\mathcal{O}(2^{|\Omega|}T)$ \cite{ramasso2007reconnaissance}. As a recommendation, when $|\Omega|$ is large, it is preferable to avoid matrix calculus and use the binary format as means to encode focal elements \cite{haenni2003implementing},   
    \item[--] Finally, it is assumed that sensors required for measuring the effectiveness are available. First, it might not be the case and one needs to assess cost/benefits of adding the required sensors. Second, it might be the case but with sensors not as relevant as desired (for instance, a luminosity sensor is made available but not exactly at the desired location). In that case, one can determine some contextual discounting factors \cite{mercier2012belief}, but the challenge remains on the determination of the discounting factor values.
\end{itemize} 
\newpage
\section{Acknowledgment}
\noindent This work has been supported by GFI Informatique, Innovation group, Saint-Ouen, France. The authors wish to thank the reviewers for their valuable suggestions that greatly helped improve the quality of this paper.\\
\noindent The authors are also thankful to Emmanuel Ramasso (emmanuel.ramasso@femto-st.fr) for his help in better understanding some specific points of the TBM theory.
\medskip

\bibliographystyle{elsarticle-num}
\bibliography{bibliography}

\end{document}